\documentclass{article}

\PassOptionsToPackage{numbers}{natbib}
\usepackage{ai_lab_2026}

\usepackage[utf8]{inputenc} 
\usepackage[T1]{fontenc}    
\usepackage[pagebackref=true,breaklinks=true,letterpaper=true,colorlinks,bookmarks=false]{hyperref}
\usepackage{url}            
\usepackage{booktabs}       
\usepackage{amsfonts}       
\usepackage{nicefrac}       
\usepackage{microtype}      
\usepackage{xcolor}         
\usepackage{amsmath}
\usepackage{multirow,multicol}
\usepackage{graphicx}
\usepackage{makecell}
\usepackage{subcaption}
\usepackage{wrapfig}        
\usepackage{float}         

\usepackage[table]{xcolor}
\definecolor{ngreen}{HTML}{03C75A}
\definecolor{ngreenlight}{HTML}{E8F8EF}
\definecolor{citegreen}{HTML}{1B7A3E}
\hypersetup{citecolor=citegreen}

\usepackage{tcolorbox}
\usepackage{microtype}
\usepackage{cancel}

\usepackage{xspace}

\newcommand{\method}{{\texttt{WATCH}}\xspace}
\newcommand{\rev}[1]{#1}

\usepackage{pifont}         
\newcommand{\cmark}{\ding{51}}
\newcommand{\xmark}{\ding{55}}

\usepackage{xspace}
\usepackage[font=small]{subcaption}

\usepackage{graphicx} 
\usepackage[dvipsnames]{xcolor}

\usepackage{lipsum}
\usepackage{multirow}
\usepackage{multicol}
\usepackage{soul}
\IfFileExists{latexml.sty}{\usepackage{latexml}}{\newif\iflatexml\latexmlfalse}

\makeatletter
\iflatexml\renewcommand{\@logotitlebar}{}\fi
\makeatother
\title{Learning Transferable World-Action Models from Task-Paired Human Videos}

\iflatexml
\author{
    Jeongeun Park* \\ NAVER AI Lab, Georgia Institute of Technology
    \and
    Taekyung Kim \\ NAVER AI Lab
    \and
    Dongyoon Han \\ NAVER AI Lab
    \and
    Sangdoo Yun* \\ NAVER AI Lab
}
\date{*Corresponding authors \quad \url{https://watch-wam.github.io}}
\else
\author{
    Jeongeun Park$^{1,2*}$
    \quad
    Taekyung Kim$^{1}$
    \quad
    Dongyoon Han$^{1}$
    \quad
    Sangdoo Yun$^{1*}$
    \\[2mm]
    $^{1}$NAVER AI Lab
    \quad
    $^{2}$Georgia Institute of Technology
    \\ 
    \small{$^*$ corresponding authors}    
    \\[2mm]
    \url{https://watch-wam.github.io}     
}
\fi

\begin{document}

\maketitle
\begin{abstract}
World-action models inherit broad visual and physical priors from video pre-training. However, standard within-video next-frame prediction allows them to rely on appearance shared between the context and future, rather than learning interactions that transfer across scenes and embodiments. 
We introduce \method, a cross-video pre-training that learns from task-paired human demonstrations collected across diverse environments.
Conditioned on the video and actions of one demonstration, the model jointly predicts another demonstration’s actions and future frames from its current observation.
Predicting one demonstration from another encourages the model to learn task-relevant interactions that transfer across scenes and demonstrators. 
Our pre-training shows that these capabilities can be learned from pairs of egocentric human videos alone, without requiring matched human–robot demonstrations.
Analyses before robot post-training show improved motion alignment across embodiments and better action and video prediction on unseen robots, with representations closer to those learned from robot pairs. 
After robot post-training, \method improves MimicDroid in-context success from $40.9\%$ to $68.3\%$ with Cosmos Policy and from $59.8\%$ to $77.7\%$ with Cosmos3, with the largest gains on the unseen levels.
Gains persist under context-free post-training across robot embodiments.
\method outperforms within-video pre-training and co-training baselines,
supporting cross-video prediction as a scalable approach to learning transferable world-action representations.

\end{abstract}

\section{Introduction}
\label{sec:intro}

Video models learn rich spatiotemporal representations by predicting how visual scenes evolve, which increasingly serve as world models~\citep{nvidia2025cosmos, wan2025} and backbones for robot foundation models~\citep{kim2026cosmospolicy, yuan2026fastwam, ye2026world}. 
World-action models extend this predictive capability to robot control by jointly modeling actions and their visual consequences, providing a learning signal not only for which action to take but also for how that action changes the world. 
These models typically build on video backbones pretrained with next-frame prediction over large-scale datasets spanning diverse environments and embodiments~\citep{nvidia2025cosmos, assran2025vjepa2}. Ideally, such pre-training should learn action representations that preserve the motion and interaction shared across different scenes and embodiments~\citep{zheng2026egoscale, ye2025lapa}, rather than the appearance and execution style specific to a particular environment or demonstrator.

However, standard video pre-training predicts future frames \textbf{within a single demonstration}~\citep{nvidia2025cosmos, assran2025vjepa2}, where the context and target share scene appearance, viewpoint, and collection source.
This continuity allows prediction to exploit demonstration-specific visual cues without requiring transferable interactions to be separated from those factors.
Such entanglement can make post-training for robot control more difficult, as the model must learn to predict robot actions under differences in environment, embodiment, and viewpoint from the pre-training data.
Even with diverse training data, the within-video objective does not explicitly encourage the model to represent similar motions and interactions consistently across different scenes and embodiments.


A straightforward way to encourage such consistency is to condition prediction on a different demonstration of the same task.
Under this \textbf{cross-video training}, the reference supplies the task and its motion, while the query's current observation provides the context for predicting its actions and future frames.
Because the two demonstrations differ in scene, viewpoint, or demonstrator, relating the reference to the query encourages the model to capture shared motion and interaction beyond demonstration-specific visual cues.
Related cross-embodiment policies learn such reference-to-query correspondences for robot control using task-matched human--robot demonstrations~\citep{jain2024vid2robot, shah2025mimicdroid, zhou2026zero, park2026retrieve}.
Using this pairing strategy for large-scale pre-training, however, would require corresponding robot demonstrations, whose collection incurs the costs of robot hardware and teleoperation.
\rev{These policies also need a demonstration at test time, whereas \method uses cross-demonstration prediction only for pre-training, so its prior also benefits policies without one.}



\begin{figure}[!t]
    \centering
    \includegraphics[width=0.995\linewidth]{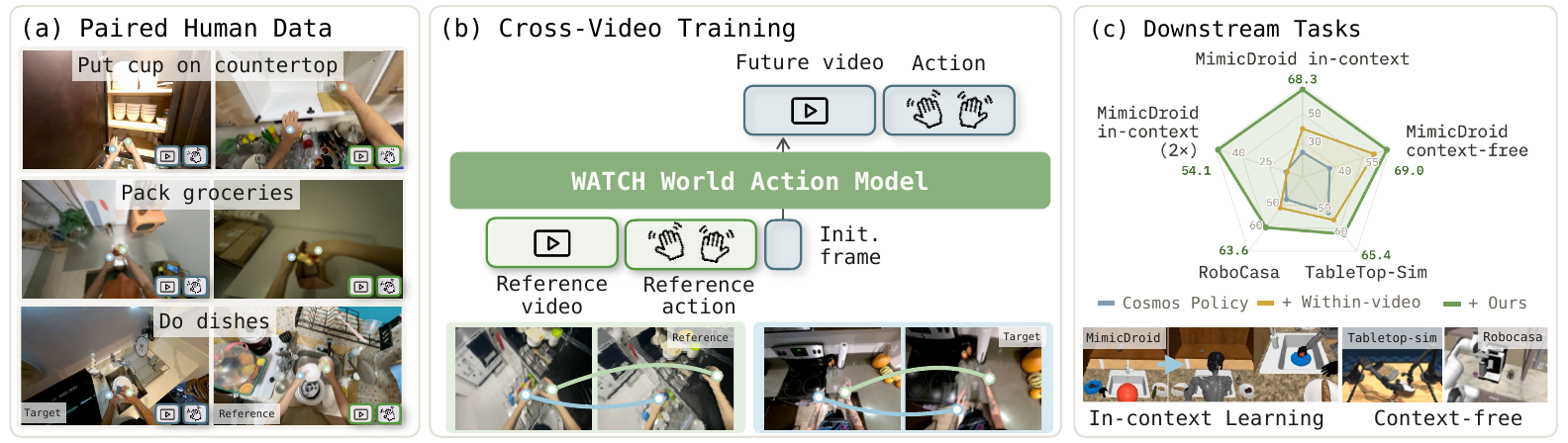}
    \caption{\textbf{\method overview.} 
    \textbf{(a)} 
    Human videos of the same task are paired across collection sources and scenes.
    \textbf{(b)} 
    Conditioned on the reference video and actions, the model predicts the target's actions and future frames from its initial frame.
    The reference's appearance does not transfer, so what it can pass on is the interaction. 
    \textbf{(c)} 
    After robot post-training, \method substantially improves success rates over the base model in both in-context and context-free settings.
    }
    \label{fig:teaser}
\end{figure}

\rev{Egocentric human} demonstrations can be collected across people and everyday environments without robot hardware or teleoperation~\citep{yang2025egovla, hoque2026egodex, punamiya2026egoverse, zheng2026egoscale}, offering a potentially more scalable and cost-effective source of pre-training data. 
To learn from this diversity through cross-video pre-training without matched robot demonstrations, we introduce \textbf{\method}, \textbf{W}orld-\textbf{A}ction model \textbf{T}raining from \textbf{C}ross-\textbf{H}uman videos. 
Our key insight is that demonstrations of the same task from different collection sources~\citep{punamiya2026egoverse, hoque2026egodex} already vary in demonstrator, viewpoint, and environment, and can therefore serve as cross-embodiment pairs without matched robot data.
This formulation allows pre-training to scale with \rev{egocentric human} demonstrations alone, reserving costly robot data collection for subsequent adaptation to the target embodiment.
As shown in Fig.~\ref{fig:teaser}, the model conditions on one human demonstration and predicts the actions and future frames of another, improving downstream robot success in both in-context and context-free settings after robot post-training.

Our analyses (\S\ref{sec:what-learns}) suggest that cross-video pre-training learns more embodiment-invariant features through joint action--video supervision, aligning motions across collection sources while preserving motion information and focusing attention on hand--object interactions.
These changes accompany improved robot post-training (\S\ref{sec:experiments}): MimicDroid~\citep{shah2025mimicdroid} success rises from $40.9\%$ to $68.3\%$ with Cosmos Policy, with gains also observed across backbones and embodiments under context-free post-training.
Comparisons with within-video learning~\citep{yang2025egovla,luo2025being}
and co-training~\citep{punamiya2026egoverse,zheng2026egoscale}, together with improvements from more paired human data, support cross-demonstration prediction as a scalable route to transferable
world-action representations.


Our main contributions are as follows: 
(1) We introduce \method, a cross-video world-action pre-training objective that conditions on one human demonstration while predicting another, together with task-matched human-to-human pairs that supply embodiment and environment variation without paired robot demonstrations.
(2) We show that cross-video pre-training shifts representations from source-specific appearance toward task-relevant motion and interaction (\S\ref{sec:what-learns}).
(3) We show consistent gains across video-model backbones, post-training protocols, and evaluation environments, largest on the unseen levels (\S\ref{sec:experiments}).


    
\section{\method: World-Action Model Training from Cross-Human Videos}
\label{sec:method}
We introduce \method, a human-data-only pre-training stage that aims to learn embodiment-invariant interaction representations before robot post-training.
We first review within-video world-action modeling (\S\ref{sec:prelim}) and what the current video prior may miss (\S\ref{sec:motivations}), then describe how task-paired human demonstrations and cross-video action--video prediction encourage the model to capture interactions shared across scenes and embodiments, and how the resulting model supports both in-context and context-free robot post-training (\S\ref{sec:cross_video}).
We provide a detailed review of robot policy pre-training and learning from human demonstrations in Appendix~\ref{app:related}.

\subsection{Preliminary}
\label{sec:prelim}

\noindent\textbf{World-action models and within-video training.}
World-action models turn the spatiotemporal priors of video models into policies by jointly predicting actions and their visual outcomes \citep{li2025unified,zhu2025uwm,kim2026cosmospolicy}.
Cosmos Policy~\citep{kim2026cosmospolicy} implements this principle by incorporating state and action representations into the latent token sequence of a pre-trained video model; we follow its recipe on Cosmos-Predict2.5~\citep{nvidia2025cosmos} (\S\ref{app:training}).
The latent sequence is split into clean conditioning slots and target slots, the latter holding future frames and actions.
Training follows the backbone's rectified-flow objective,
$\mathcal{L}(\theta) = \mathcal{L}_{\mathrm{visual}}(\theta) + \lambda\,\mathcal{L}_{\mathrm{action}}(\theta)$,
where each term applies velocity matching to the corresponding target slots, with no loss on conditioning slots.
These models typically follow a \textbf{within-video} formulation: conditioning observations and prediction targets come from the same demonstration or episode.
Writing $I_t$ for the frame, $s_t$ for the proprioceptive state, $\ell$ for language instruction, and $a_t$ for the action at time $t$, the model learns
$p_{\theta}(a_{t:t+C}, I_{t+1:t+C+1} \mid I_t, s_t, \ell)$
over a chunk size of $C$ steps, where $x_{i:j}$ denotes $x_i, \ldots, x_{j-1}$, so that $a_{t:t+C}$ holds $C$ actions.
The condition and target therefore share the same scene, viewpoint, and demonstrator (or embodiment).

\subsection{Motivation}\label{sec:motivations}

\begin{figure}[t]
    \centering
    \begin{subfigure}[t]{0.33\textwidth}
        \includegraphics[width=0.964\linewidth]{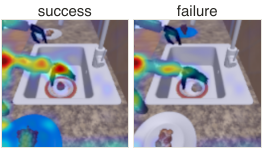}
        \caption{Post-trained original model}
    \end{subfigure}\hfill
    \begin{subfigure}[t]{0.33\textwidth}
        \includegraphics[width=0.964\linewidth]{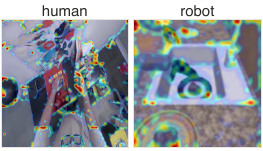}
        \caption{Original video model}
    \end{subfigure}\hfill
    \begin{subfigure}[t]{0.33\textwidth}
        \includegraphics[width=0.964\linewidth]{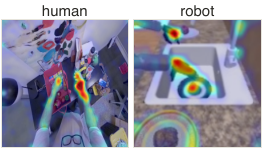}
        \caption{Cross-video pre-training (ours)}
    \end{subfigure}
    \caption{\textbf{Where a world-action model looks.} 
    Attention to the current frame, grouped by model in the order the text reads them.
    (a) 
    After robot post-training of the baseline video backbone (Cosmos-Predict2.5), attention focuses on the hand and object in a successful rollout and on the forearm in a failed one ({\small\texttt{PnPSinkToCabinet}}).
    (b) 
    The original video model attends broadly to edges and background on both robot frames and human demonstrations, which offer a scalable source of interaction and motion cues for robot learning.
    (c) 
    On the same frames, our cross-video pre-training shifts attention toward the acting hand and manipulated object.
    }
    \vspace{-10pt}
    \label{fig:motivation}
\end{figure}

\noindent\textbf{What the video prior may miss.}
We examine Cosmos Policy~\citep{kim2026cosmospolicy} obtained by directly post-training Cosmos-Predict2.5~\citep{nvidia2025cosmos} on robot data.
In the rollouts of Fig.~\ref{fig:motivation}a, attention focuses on the hand near the manipulated object in the successful one, but on the wrist and forearm in the failed one.
To assess whether the original video backbone emphasizes these cues, we inspect its attention on human demonstrations as well as robot observations.
In both cases, attention is broadly distributed over edges and background rather than concentrated on the interacting hand and object (Fig.~\ref{fig:motivation}b).
We hypothesize that this partly reflects its pre-training objective, which models visual change throughout the scene without prioritizing the interaction relevant for control.
This motivates \textbf{cross-video pre-training}, where another demonstration of the same task guides prediction in the current scene: predicting across demonstrations with different object placements and demonstrators encourages the model to capture the shared interaction while adapting the motion to the current hand--object configuration.
Consistent with this, cross-video pre-training shifts attention toward the acting hand and manipulated object in both human and robot observations (Fig.~\ref{fig:motivation}c).

\subsection{Cross-Video World-Action Modeling}
\label{sec:cross_video}

\begin{figure}
    \centering
    \includegraphics[width=0.99\linewidth]{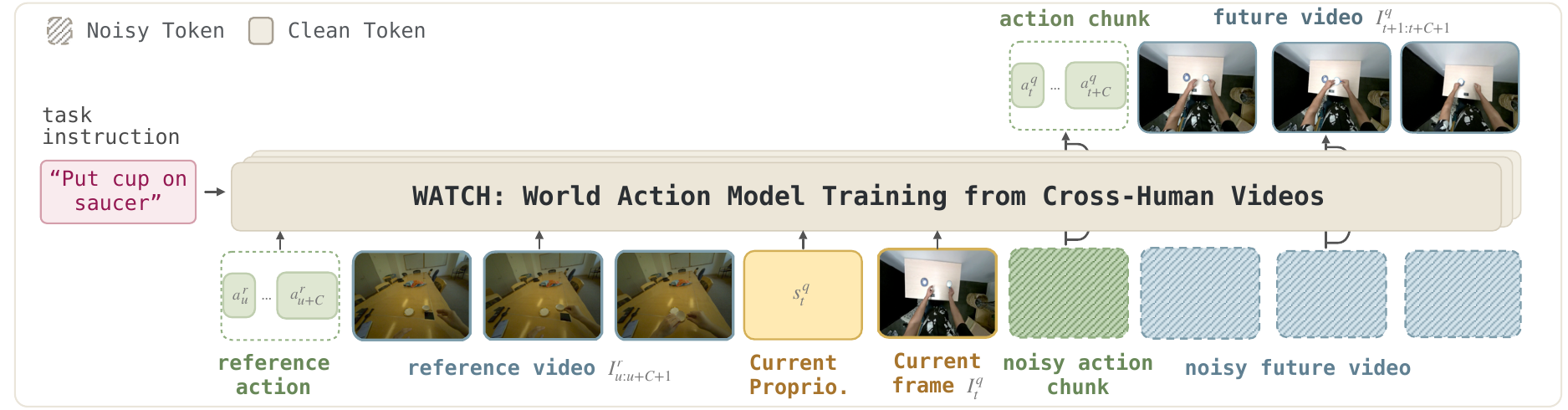}
    \caption{\textbf{Overview of \method.} The reference video and its action
    trajectory, from a different demonstration of the same task, enter as clean
    conditioning tokens together with the query's current frame and
    proprioceptive state; the query's action chunk and future frames are the
    noised tokens the model denoises. Every token is tagged with its slot type and
    physical time, so the same sequence layout carries over to post-training
    with a human demonstration as the reference or with the reference slots
    removed. The loss is the joint velocity-matching objective of
    \S\ref{sec:prelim}, applied to the generated tokens.}
    \label{fig:model}
    \vspace{-1em}
\end{figure}
 
\noindent\textbf{Human data and cross-video pairs.}
Human demonstrations collected from different people, with different recording rigs and in different environments, vary along the axes on which a robot differs from a human, and task-matched pairs across these sources supply this variation without requiring matched robot data.
We construct query–reference pairs from different episodes of EgoVerse~\citep{punamiya2026egoverse}.
Human actions are derived from subsequent body configurations and expressed with states in a shared human–robot task space, using a canonical frame fixed at each episode's start.
A semantic task taxonomy restricts candidates to compatible activities, which we rank by dynamic-time-warping distance between motion trajectories.
Motion descriptors are standardized within each window, which reduces their sensitivity to motion magnitude so that interactions can match despite differences in reach.
Each query's reference set $\mathcal{R}(q)$ holds up to $8$ nearest matches from its own source and up to $7$ from other sources, the latter admitted within a fixed factor of the best score.
Training samples one reference uniformly from this set at each step, from any source (\S\ref{app:pairs}).


This procedure yields 10.99M query--reference pairs, $10.0$K hours of paired $3$-second windows counted over all pairs.
Each query has 10.52 references on average, including 6.35 cross-source matches; 90.7\% of queries have at least one cross-source reference.
Compared with same-task random cross-source pairs, the selected matches show higher trajectory-shape cosine similarity by 0.36, indicating agreement in motion beyond task semantics (\S\ref{app:stats}).

\noindent\textbf{Reference-conditioned joint prediction.}
As illustrated in Fig.~\ref{fig:model}, the model receives the current query
frame and state together with the reference video and actions. For a query
window $q$ and a reference window $r$ drawn from its reference set, we model
\begin{equation}
p_{\theta}\left(
a^q_{t:t+C},\, I^q_{t+1:t+C+1}
\mid
I^q_t,\, s^q_{t},\, \ell,\,
I^r_{u:u+C+1},\, a^r_{u:u+C}
\right),
\qquad r \sim \mathcal{R}(q),
\label{eq:cross}
\end{equation}
over a prediction horizon $C$, where $u$ is the start of the reference window and the task description $\ell$ enters as text on both backbones.
The reference supplies an example of task progression while the query observation specifies the target scene and source, so predicting the query future requires transferring the demonstrated interaction into the query's visual and geometric context.
Inputs and outputs are encoded as six slots: current query frame ($I^q_t$), query state ($s^q_{t}$), reference video ($I^r_{u:u+C+1}$), reference actions ($a^r_{u:u+C}$), predicted future video ($I^q_{t+1:t+C+1}$) and actions ($a^q_{t:t+C}$). Each carries its own positional encoding over slot type and physical time, so where the reference begins and where the action begins are marked by slot rather than by absolute offset, and the encoding carries over to post-training as slots are repopulated or dropped.
Following the joint video-action prediction recipe of Cosmos Policy, we implement \method on Cosmos-Predict2.5 and Cosmos3~\citep{nvidia2026cosmos3} using the joint objective of \S\ref{sec:prelim}, applied only to the generated slots, which ties the predicted motion to its expected visual effect while retaining the backbone's dynamics prior.
A query and its reference are $3$-second windows, the action term is weighted by $\lambda = 48$, and post-training on the target robot keeps the same objective and slot layout (\S\ref{app:architecture}).

\noindent\textbf{Post-training with and without context.}
The model trained by \method supports both in-context and context-free post-training, with predicted actions expressed in the target robot's native control space.
In-context post-training conditions robot action and future-video prediction on a human reference demonstration.
Context-free post-training removes the reference slots and predicts robot actions and future frames from the robot's current observation, state, and task instruction.
Both protocols retain the same backbone and joint prediction objective.
\section{What Cross-Video Pre-Training Learns}
\label{sec:what-learns}
Motivated by the hypothesis that cross-video prediction encourages the model to capture interactions shared across demonstrations, we examine what \method learns before robot post-training: how it uses task information from a reference to predict interactions in another demonstration, through reference sensitivity, motion alignment, and hand--object attention.
We then assess how this learning extends to unseen human and robot demonstrations, including whether human pairs yield representations similar to those learned from robot pairs.
We use held-out human pairs from EgoVerse~\citep{punamiya2026egoverse}, paired robot data from RoboMIND~\citep{wu2025robomind}, and robot observations from MimicDroid~\citep{shah2025mimicdroid}, the post-training benchmark of \S\ref{sec:setup}.
The results suggest that cross-video pre-training captures task-relevant motion and interactions across embodiments.

\begin{figure}[t]
    \centering
    \begin{subfigure}[t]{0.364\textwidth}
        \includegraphics[width=\linewidth]{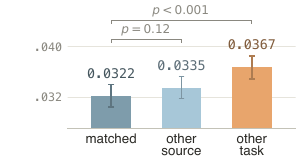}
        \caption{Sensitivity to the reference}
    \end{subfigure}
    \hfill
    \begin{subfigure}[t]{0.62\textwidth}
        \includegraphics[width=\linewidth]{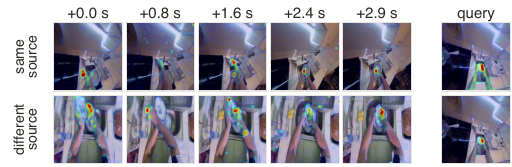}
        \caption{Action-token attention}
    \end{subfigure}
    \caption{
    \textbf{Sensitivity to the reference demonstration.}
    \textbf{(a)} Action MSE (lower is better) over $137$ held-out episodes using a matched reference, a similar-motion reference from a different collection source, or a different-task reference from the same source.
    Brackets: Holm-corrected $p$-values, episode-paired Wilcoxon test.
    Our model is less sensitive to a change of source than of task, suggesting that our training encourages a focus on task-relevant actions. 
    \textbf{(b)}  Action-token attention for the same query (do dishes) with references from the same collection source (top) and a different one (bottom).
    Even with a reference from a different source, attention concentrates on interactions central to task execution.
    }
    \label{fig:robustness}
\end{figure}

\begin{figure}[t]
    \centering
    \includegraphics[width=\linewidth]{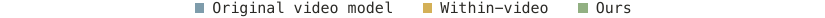}\\[0pt]
    \begin{subfigure}[t]{0.333\textwidth}
        \includegraphics[width=\linewidth]{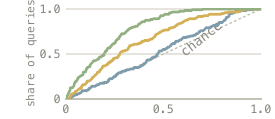}
        \caption{Retrieval, human}
    \end{subfigure}\hfill
    \begin{subfigure}[t]{0.284\textwidth}
        \includegraphics[width=\linewidth]{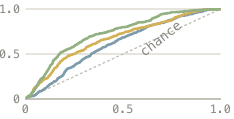}
        \caption{Retrieval, robot}
    \end{subfigure}\hfill
    \begin{subfigure}[t]{0.351\textwidth}
        \includegraphics[width=\linewidth]{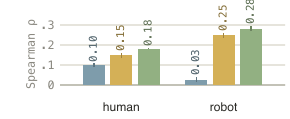}
        \caption{Held-out embodiment regression}
    \end{subfigure}
    \caption{
    \textbf{Representation alignment and motion transfer across embodiments.}
    Action-slot features from layers $10$--$27$, with retrieval ranks and regression scores averaged over these layers, on $359$ held-out EgoVerse windows (human) and $396$ RoboMIND windows from three robots (robot).
    \textbf{(a, b)} Motion retrieval: share of queries whose matched window from a different collection source (a) or robot (b) falls within a normalized rank; a higher curve is better, the diagonal is chance.
    \textbf{(c)} Held-out embodiment regression: a regressor fit on two embodiments predicts nine future-motion quantities on the third; bars are the mean Spearman correlation over quantities and folds, error bars the standard error over layers (\S\ref{app:repr}).
    }
    \label{fig:invariance}
    \vspace{-1.6em}
\end{figure}

\noindent\textbf{Conditioning is more sensitive to task changes.}
\label{sec:ana_sensitivity}
We compare matched references with similar-motion references from a different collection source and different-task references from the same source in Fig.~\ref{fig:robustness}a.
Action MSE increases from $0.0322$ to $0.0335$ under the source change and to $0.0367$ under the task change, only the latter significant after Holm correction.
In the visualized examples, attention remains concentrated on the acting hands and manipulated object when the reference source changes (Fig.~\ref{fig:robustness}b; \S\ref{app:cswap}).
Together, these observations suggest that reference conditioning tracks task-relevant interactions across collection sources.


\noindent\textbf{Embodiment-invariant motion representations from human demonstrations.}
\label{sec:invariance}
Cross-embodiment motion transfer requires both alignment of corresponding movements and preservation of information about how those movements unfold.
We examine both properties using action-slot features of the original video model, the within-video baseline, and \method on $359$ held-out EgoVerse windows and $396$ RoboMIND windows from three robots (Fig.~\ref{fig:invariance}).
In motion retrieval (Fig.~\ref{fig:invariance}a,b), \method ranks matching motions from different human sources and different robots earlier than both baselines.
To test whether this alignment also preserves transferable motion information, we fit a regressor on two embodiments and evaluate it on the third (Fig.~\ref{fig:invariance}c).
It predicts future motion from \method's features at a Spearman correlation of $0.18$ on human and \rev{$0.28$} on robot windows, against \rev{$0.10$ and $0.03$ for the original video model and $0.15$ and $0.25$ for the within-video baseline, all models receiving the same reference input}.
These complementary results suggest that human cross-video pre-training supports both cross-embodiment correspondence and a transferable encoding of motion, extending to unseen robots.


\begin{figure}[t]
    \centering
    \vspace{-8pt}
    \begin{subfigure}[t]{0.495\textwidth}
        \includegraphics[width=\linewidth]{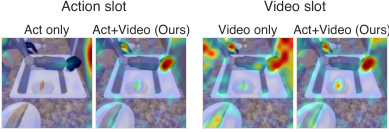}
        \caption{\texttt{TurnOnFaucet}}
    \end{subfigure}\hfill
    \begin{subfigure}[t]{0.495\textwidth}
        \includegraphics[width=\linewidth]{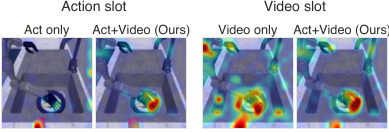}
        \caption{\texttt{PnPSinkToRightCounterPlate}}
    \end{subfigure}
    \caption{
    \textbf{Attention visualizations under different pre-training supervision.} 
    All models use cross-video pairing and differ in which reference modalities they receive and predict: actions only (Act only), future frames only (Video only), or both (Act+Video, ours).
    Joint supervision concentrates attention on the acting hand and the manipulated object on both the action and the video slot.
}
    \vspace{-20pt}
    \label{fig:attention}
\end{figure}

\noindent\textbf{Joint supervision focuses attention on interactions.}
\label{sec:anal_mod}
To understand the role of predicting both motion and its visual consequences, we compare attention under action-only, video-only, and joint training with the same cross-video pairing (\S\ref{sec:cross_video}), using the respective modalities as reference inputs and prediction targets.
Fig.~\ref{fig:attention} shows attention from the corresponding output slots on identical query frames.
Action-only attention peaks near image boundaries, away from the manipulated object, while video-only attention spreads across the interaction region and surrounding scene.
Joint training concentrates both action-slot and video-slot attention on the acting hand and object, around the faucet in {\small\texttt{TurnOnFaucet}} and the grasp in {\small\texttt{PnPSinkToRightCounterPlate}}.
This shared focus suggests that jointly predicting motion and its visual consequences encourages both outputs to attend to the interaction connecting them.

\newcommand{\offlineTableBody}{%
\footnotesize
\setlength{\tabcolsep}{1.8pt}
\renewcommand{\arraystretch}{1.75}
\captionof{table}{
\textbf{Evaluation on held-out robot data.}
End-effector action and video prediction on RoboMIND, unseen by every model; \emph{original model} is the video model before any action training. Other sets are in \S\ref{app:heldout-more}.
}
\label{tab:offline}
\begin{tabular}{l|cc|cc}
    \toprule
    & \multicolumn{2}{c|}{Action} & \multicolumn{2}{c}{Generated video} \\
    \cmidrule(lr){2-3} \cmidrule(lr){4-5}
    & MSE $\downarrow$ & acc@$0.1$ $\uparrow$ & FVD $\downarrow$ & FID $\downarrow$ \\
    \midrule
    Original model & -- & -- & 290.7 & 12.48 \\
    Within-video   & 0.046 & 55.6 & 1485.1 & 53.53 \\
    \rowcolor{ngreenlight}
    \method (Ours) & \textbf{0.039} & \textbf{68.5} & \textbf{154.9} & \textbf{8.33} \\
    \bottomrule
\end{tabular}}
\newcommand{\matchFigureBody}{%
\includegraphics[width=\linewidth]{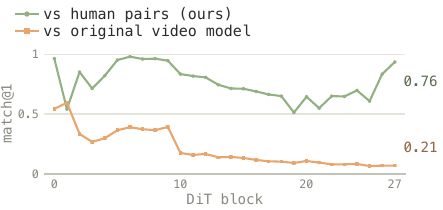}
\captionof{figure}{\textbf{Alignment with a robot-trained model.}
Match@$1$ for human-trained \method (green) and the original video model (orange), measured against the robot-trained model on $359$ EgoVerse windows (robot windows: Fig.~\ref{fig:match-layers}). 
Numbers at right: mean over all layers.
}
\label{fig:tsne-pooled}}
\iflatexml
\begin{table}[t]
\centering
\offlineTableBody
\end{table}
\begin{figure}[t]
\centering
\matchFigureBody
\end{figure}
\else
\begin{figure}[t]
\centering
\vspace{14pt}
\begin{minipage}[t]{0.47\textwidth}
\vspace{0pt}
\centering
\offlineTableBody
\end{minipage}\hfill
\begin{minipage}[t]{0.50\textwidth}
\vspace{0pt}
\centering
\matchFigureBody
\end{minipage}
\vspace{-10pt}
\end{figure}
\fi

\noindent\textbf{The prior generalizes to unseen robot data.}
We evaluate whether a model pre-trained on human demonstrations generalizes to robot data unseen during pre-training, using action and video prediction on RoboMIND~\citep{wu2025robomind} (Table~\ref{tab:offline}).
Compared with the within-video baseline using query context alone, reference-conditioned \method lowers action MSE from $0.046$ to $0.039$ and raises accuracy from $55.6\%$ to $68.5\%$.
Video FVD and FID also improve over the original video model, whereas within-video pre-training worsens both.
These results support generalization from human demonstrations to unseen robots in both action and video prediction.


\noindent\textbf{Human and robot pairs induce similar representations.}
We examine whether training on human demonstrations yields robot-motion representations similar to those learned directly from robot demonstrations.
We compare \method trained on $1$K hours of human pairs with a counterpart trained on $1$K hours of RoboMIND~\citep{wu2025robomind} robot pairs using the same cross-video objective (\S\ref{app:robomind-setup}).
Both are evaluated on the same $359$ EgoVerse and $300$ RoboMIND windows, the latter seen during pre-training only by the robot-trained model.
We align the human-trained and original video models' action-slot features to the robot-trained model's features on half of the windows, then measure cross-model window matching on the rest (\S\ref{app:robomind-mid}).
Human-trained \method achieves a mean match@$1$ of $0.76$ on human and $0.60$ on robot windows over all layers, compared with $0.21$ and $0.19$ for the original video model (Fig.~\ref{fig:tsne-pooled}).
This correspondence supports human demonstrations as a scalable source for learning robot-relevant motion representations.
\section{Post-Training Experiments}
\label{sec:experiments}

We evaluate whether the representation properties identified in our analyses translate into improved robot control after post-training on robot demonstrations.
\S\ref{sec:prior} compares \method with prior policies under in-context and context-free settings.
\S\ref{sec:ablation} examines how the pre-training objective, reference set, supervised modalities, and amount of human video affect downstream performance.

\subsection{Experimental Setup}
\label{sec:setup}
\mbox{}\par\nobreak\vspace{-2\baselineskip}
\begin{wraptable}{r}{0.47\textwidth}
\small
\vspace{-0.9em}
\centering
\caption{\textbf{In-context post-training on MimicDroid.} Success rate (\%);
\textit{Avg.} over L1--L3. $\dagger$: reported by prior work under its own evaluation data. 
}
\label{tab:icl-main}
\vspace{-7pt}
\newcommand{\grouprow}[1]{\multicolumn{5}{l}{\rule{0pt}{6.5pt}\textcolor{gray}{\scriptsize\textit{#1}}} \\[-1.5pt]}
\small
\setlength{\tabcolsep}{4pt}
\renewcommand{\arraystretch}{0.98}
\begin{tabular}{l|cccc}
    \toprule
    Method & L1 & L2 & L3 & Avg. \\
    \midrule
    \grouprow{ICL baselines}
    Vid2Robot~\citep{jain2024vid2robot} & \underline{90.8} & 51.0 & 0.8 & 47.6 \\
    HAND$^{\dagger}$~\citep{hong2025hand} & 29.0 & 21.0 & 1.0 & 17.0 \\
    MimicDroid$^{\dagger}$~\citep{shah2025mimicdroid} & 59.0 & 44.0 & 26.0 & 43.0 \\
    \grouprow{VLA baselines}
    $\pi_{0.5}$~\citep{pi05} & 79.2 & 38.3 & 8.3 & 41.9 \\
    GR00T-N1.7~\citep{nvidia2025gr00t} & 78.3 & 59.2 & 10.0 & 49.2 \\
    \midrule
    \grouprow{World-action models}
    Cosmos Policy~\citep{kim2026cosmospolicy} & 75.0 & 29.4 & 18.3 & 40.9 \\
    \rowcolor{ngreenlight}
    \quad + \method (Ours) & \textbf{91.7} & 72.5 & \underline{40.8} & \underline{68.3} \\
    \addlinespace[1pt]
    Cosmos3~\citep{nvidia2026cosmos3} & 69.2 & \underline{79.4} & 30.8 & 59.8 \\
    \rowcolor{ngreenlight}
    \quad + \method (Ours) & \textbf{91.7} & \textbf{86.5} & \textbf{55.0} & \textbf{77.7} \\
    \bottomrule
\end{tabular}
\end{wraptable}

\noindent\textbf{In-context benchmarks.}
In-context learning from reference demonstrations is an emerging direction in robot learning \citep{generalist2026gen15,skild2026s1}.
We evaluate whether task-paired human pre-training improves in-context control on MimicDroid~\citep{shah2025mimicdroid} with the GR1 humanoid.
The benchmark ships human demonstrations but no robot-executed trajectories or evaluation scenes: MimicGen~\citep{mandlekar2023mimicgen} scene draws and retargeting turn its three demonstrations per task into our $253$ GR1 training episodes, the $100$ reference demonstrations per task and the evaluation scenes (\S\ref{app:mimicdroid}).
We follow the MimicDroid L1--L3 split: L1 is the training tasks in the
training scenes, L2 changes appearance and objects within the layout, and L3
holds out the layout; each episode is conditioned on a demonstration from its
test setting. To probe how the policy uses the reference, we add an
infeasible-demonstration setting: since human demonstrations are often much
faster than the robot can execute, we replay the reference demonstrations at $2\times$
speed during both post-training and evaluation, faster than the robot can
follow.

\noindent\textbf{Context-free benchmarks.}
To test whether the representations remain useful without a reference and
transfer across robots, we evaluate three settings with no reference during
post-training or evaluation: context-free MimicDroid on the GR1 humanoid,
with L1 and L2 reported as \emph{Easy} and \emph{Hard};
TableTop-Sim~\citep{im2026twinvla} with bimanual ALOHA; and
RoboCasa~\citep{nasiriany2024robocasa} Kitchen-H50 with a single-arm Franka
and $50$ human demonstrations per task. For RoboCasa only, Cosmos3
pre-training additionally includes RoboMIND data (\S\ref{app:training}).

\noindent\textbf{Backbones and baselines.}
We use Cosmos Policy~\citep{kim2026cosmospolicy} on the $2$B
Cosmos-Predict2.5~\citep{nvidia2025cosmos} backbone, with the post-training
objective reduced to future-video and action prediction
(\S\ref{app:training}).
We also apply \method to Cosmos3 Edge-2B~\citep{nvidia2026cosmos3}, a
backbone already trained on human and robot action data under a within-video
objective. In-context baselines are Vid2Robot~\citep{jain2024vid2robot},
HAND~\citep{hong2025hand}, MimicDroid~\citep{shah2025mimicdroid},
$\pi_{0.5}$~\citep{pi05}, and GR00T-N1.7~\citep{nvidia2025gr00t}; the
context-free comparison adds
MolmoAct2~\citep{fang2026molmoact2actionreasoningmodels}. All methods we
train use the same robot demonstrations as \method; $\dagger$ marks numbers
reported by prior work under its own evaluation data.

\subsection{Comparison with Prior Methods}
\label{sec:prior}
\noindent\textbf{In-context post-training.}
Table~\ref{tab:icl-main} compares \method with prior  demonstration-conditioned
policies~\citep{jain2024vid2robot,hong2025hand,shah2025mimicdroid} and
vision-language-action policies~\citep{pi05,nvidia2025gr00t} under the MimicDroid L1--L3 split.
With Cosmos Policy~\citep{kim2026cosmospolicy}, \method raises the average
success rate from $40.9\%$ to $68.3\%$, with gains at all three levels and
the largest under environment shift (L2 $29.4\%$ to $72.5\%$, L3 $18.3\%$
to $40.8\%$). The
same trend holds with
Cosmos3~\citep{nvidia2026cosmos3}, where the average increases from $59.8\%$
to $77.7\%$ and L3 from $30.8\%$ to $55.0\%$. 
Some baselines match or exceed Cosmos Policy on L1 but fall to $10\%$ or below on L3, the unseen layouts, where \method gains substantially.

\begin{table}[t]
\centering
\caption{\textbf{Context-free post-training.} Success rate (\%) without a
reference demonstration; \textit{Avg.} over the two levels of each benchmark.
}
\label{tab:bc-main}
\newcommand{\grouprowcf}[1]{\multicolumn{8}{l}{\rule{0pt}{6.5pt}\textcolor{gray}{\scriptsize\textit{#1}}} \\[-1.5pt]}
\small
\setlength{\tabcolsep}{6pt}
\renewcommand{\arraystretch}{0.98}
\begin{tabular}{l|ccc|ccc|c}
    \toprule
    \multirow{2}{*}{Method} & \multicolumn{3}{c|}{MimicDroid} & \multicolumn{3}{c|}{TableTop-Sim} & RoboCasa \\
    \cmidrule(lr){2-4} \cmidrule(lr){5-7} \cmidrule(lr){8-8}
    & Easy & Hard & Avg. & Easy & Hard & Avg. & Kitchen-H50 \\
    \midrule
    \grouprowcf{VLA baselines}
    $\pi_{0.5}$~\citep{pi05} & 78.3 & 29.8 & 54.1 & 70.4 & 34.0 & 52.2 & 54.5 \\
    GR00T-N1.7~\citep{nvidia2025gr00t} & \underline{79.2} & \underline{45.8} & 62.5 & 66.8 & 50.8 & 58.8 & 46.4 \\
    MolmoAct2~\citep{fang2026molmoact2actionreasoningmodels} & 76.7 & 28.3 & 52.5 & 66.4 & 39.6 & 53.0 & 54.3 \\
    \midrule
    \grouprowcf{World-action models}
    Cosmos Policy~\citep{kim2026cosmospolicy} & 71.7 & 26.9 & 49.3 & 69.6 & 49.6 & 59.6 & \underline{56.0} \\
    \rowcolor{ngreenlight}
    \quad + \method (Ours) & \textbf{90.8} & \textbf{47.1} & \textbf{69.0} & \underline{76.8} & 54.0 & 65.4 & \textbf{63.6} \\
    \addlinespace[1pt]
    Cosmos3~\citep{nvidia2026cosmos3} & 78.3 & 21.0 & 49.7 & 76.4 & \underline{55.2} & \underline{65.8} & 55.0 \\
    \rowcolor{ngreenlight}
    \quad + \method (Ours) & \textbf{90.8} & 36.0 & \underline{63.4} & \textbf{85.6} & \textbf{58.4} & \textbf{72.0} & 55.8 \\
    \bottomrule
\end{tabular}
\vspace{-6pt}
\end{table}

\noindent\textbf{Context-free post-training.}
Table~\ref{tab:bc-main} evaluates standard behavior-cloning policies
without reference demonstrations during either post-training or evaluation.
On MimicDroid, \method improves Cosmos Policy from $49.3\%$ to $69.0\%$
and Cosmos3 from $49.7\%$ to $63.4\%$.
The benefit extends to TableTop-Sim (bimanual ALOHA), where both backbones
improve, and to RoboCasa (single-arm Franka), where Cosmos Policy gains
$7.6$ points while Cosmos3, which already saw robot data and here RoboMIND,
gains $0.8$.
Gains with no reference during robot post-training and evaluation show that
the benefit of pre-training persists without a test-time reference.

\subsection{Ablation Studies}
\label{sec:ablation}

All variants keep the Cosmos Policy~\citep{kim2026cosmospolicy} backbone,
the robot post-training, and the human-video data fixed.
\emph{Within-video}~\citep{yang2025egovla,luo2025being} predicts the future
actions and frames of each demonstration from its own observation, without a
reference. \emph{Co-training}~\citep{punamiya2026egoverse,
zheng2026egoscale} keeps that formulation and adds robot action data from the
robot subset of EgoVerse and Humanoid Everyday~\citep{zhao2025humanoid}.
Finally, \emph{same-source top-1} conditions every query on its single
best-ranked reference from its own collection source, whereas \method (Ours)
samples from the full reference set of \S\ref{sec:cross_video}.

\begin{table}[t]
\centering
\caption{\textbf{Pre-training ablations.} Success rate (\%) on a fixed Cosmos
Policy backbone; rows vary the pre-training objective, data mixture, and
reference set (defined in the text): in-context MimicDroid with references at recorded and at $2\times$
speed, and context-free MimicDroid, TableTop-Sim, and RoboCasa Kitchen-H50.
\emph{None} is the backbone post-trained directly~\citep{kim2026cosmospolicy}.}
\label{tab:ablation}
\footnotesize
\setlength{\tabcolsep}{1.6pt}
\renewcommand{\arraystretch}{1.12}
\begin{tabular}{l|cccc|cccc|ccc|ccc|c}
    \toprule
    \multirow{3}{*}{Pre-training} & \multicolumn{8}{c|}{In-context} & \multicolumn{7}{c}{Context-free} \\
    \cmidrule(lr){2-9} \cmidrule(lr){10-16}
    & \multicolumn{4}{c|}{MimicDroid} & \multicolumn{4}{c|}{MimicDroid, $2\times$} & \multicolumn{3}{c|}{MimicDroid} & \multicolumn{3}{c|}{TableTop-Sim} & \multirow{2}{*}{\makecell{Robo-\\Casa}} \\
    \cmidrule(lr){2-5} \cmidrule(lr){6-9} \cmidrule(lr){10-12} \cmidrule(lr){13-15}
    & L1 & L2 & L3 & Avg. & L1 & L2 & L3 & Avg. & Easy & Hard & Avg. & Easy & Hard & Avg. & \\
    \midrule
    None~\citep{kim2026cosmospolicy} & 75.0 & 29.4 & 18.3 & 40.9 & 73.3 & 18.8 & 0.0 & 30.7 & 71.7 & 26.9 & 49.3 & 69.6 & 49.6 & 59.6 & 56.0 \\
    Within-video~\citep{yang2025egovla,luo2025being} & 85.8 & 44.4 & 23.3 & 51.2 & 75.0 & 7.5 & 8.3 & 30.3 & 88.3 & 40.8 & 64.6 & \underline{71.6} & 51.6 & \underline{61.6} & 58.3 \\
    Co-training~\citep{punamiya2026egoverse,zheng2026egoscale} & \underline{87.5} & \underline{58.3} & \underline{34.2} & \underline{60.0} & \textbf{88.3} & 25.2 & 5.0 & 39.5 & \underline{90.0} & \underline{42.3} & \underline{66.1} & 69.2 & \underline{52.8} & 61.0 & \underline{61.7} \\
    Same-source top-1 & 84.2 & \underline{58.3} & 33.3 & 58.6 & \underline{85.8} & \underline{26.7} & \underline{23.3} & \underline{45.3} & \underline{90.0} & 36.7 & 63.3 & 69.6 & 51.2 & 60.4 & 56.8 \\
    \rowcolor{ngreenlight}
    \method (Ours) & \textbf{91.7} & \textbf{72.5} & \textbf{40.8} & \textbf{68.3} & 84.2 & \textbf{36.5} & \textbf{41.7} & \textbf{54.1} & \textbf{90.8} & \textbf{47.1} & \textbf{69.0} & \textbf{76.8} & \textbf{54.0} & \textbf{65.4} & \textbf{63.6} \\
    \bottomrule
\end{tabular}
\end{table}

\noindent\textbf{Ablation on pre-training objective.}
\label{sec:ablation_obj}
Table~\ref{tab:ablation} shows that within-video pre-training improves
over no pre-training and co-training improves further on MimicDroid, but both leave the
representation tied to cues within individual demonstrations.
By predicting across demonstrations, \method encourages features that capture shared interactions, consistent with its larger gains on L2 and L3.
Under $2\times$-speed references, within-video pre-training loses its whole gain ($30.3\%$ against $30.7\%$ with none) whereas \method keeps most of its own, so the cross-video prior still supports learning from references the robot cannot follow.
\rev{The in-context and context-free gains are two uses of the same learned correspondence: in-context control uses it directly through the reference slots, under one reference protocol for all variants, and context-free control only through the learned features, with smaller gains.
The in-context gain is not from exposure to reference conditioning: random same-task pre-training has the same slots and cross-source references yet falls below no pre-training (Table~\ref{tab:ref-pairing}), so in-context success follows correspondence quality.
Without references, \method still beats both human-data baselines on all three context-free benchmarks, a gain from cross-demonstration prediction beyond human-data learning.}
Together, these results suggest that cross-demonstration prediction, not data diversity alone, turns diverse human experience into more embodiment-invariant representations.

\begin{wraptable}{r}{0.53\textwidth}
\vspace{-1.1em}
\centering
\caption{\rev{\textbf{Reference pairing.} Success rate (\%) on in-context
MimicDroid and context-free TableTop-Sim. X: cross-source; M: motion-matched.}}
\label{tab:ref-pairing}
\vspace{-7pt}
\scriptsize
\setlength{\tabcolsep}{2pt}
\renewcommand{\arraystretch}{1.0}
\begin{tabular}{l|cc|cccc|ccc}
    \toprule
    & & & \multicolumn{4}{c|}{MimicDroid} & \multicolumn{3}{c}{TableTop-Sim} \\
    Pre-training & X & M & L1 & L2 & L3 & Avg. & Easy & Hard & Avg. \\
    \midrule
    Random same-task & \cmark & \xmark & 75.8 & 31.5 & 9.2 & 38.8 & 70.8 & 48.8 & 59.8 \\
    Same-source top-1 & \xmark & \cmark & 84.2 & 58.3 & 33.3 & 58.6 & 69.6 & 51.2 & 60.4 \\
    \rowcolor{ngreenlight}
    \method (Ours) & \cmark & \cmark & \textbf{91.7} & \textbf{72.5} & \textbf{40.8} & \textbf{68.3} & \textbf{76.8} & \textbf{54.0} & \textbf{65.4} \\
    \bottomrule
\end{tabular}
\vspace{-1.2em}
\end{wraptable}

\noindent\textbf{\rev{Ablation on reference pairing.}}
\rev{Table~\ref{tab:ref-pairing} separates the two ingredients of the pairing.
\emph{Random same-task} draws cross-source references of the query's task at random instead of by motion matching.
It leaves no gain over no pre-training ($38.8$ against $40.9\%$ in context, $59.8$ against $59.6\%$ on TableTop-Sim; Table~\ref{tab:ablation}) and falls below within-video pre-training, since a reference that does not follow the query's motion does not make the query's future predictable.
Same-source top-1 keeps motion matching but not the other sources, and also stays well below \method.
The gain therefore needs references that share the query's motion; sampling from the full reference set, which includes other sources, adds a further gain over a single same-source match.
Task labels only gate the candidates: with half of them confused, \method retains most of its gain (App.~\ref{app:label-noise}).}

\begin{figure}[t]
    \centering
    \includegraphics[width=\linewidth]{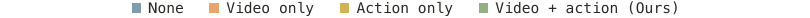}\\[0pt]
    \begin{subfigure}[t]{0.557\textwidth}
        \includegraphics[width=\linewidth]{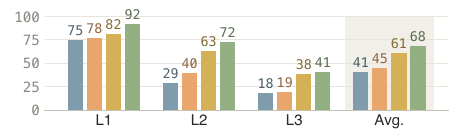}
        \caption{MimicDroid, in-context}
    \end{subfigure}\hfill
    \begin{subfigure}[t]{0.439\textwidth}
        \includegraphics[width=\linewidth]{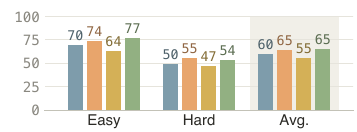}
        \caption{TableTop-Sim, context-free}
    \end{subfigure}
    \caption{\textbf{Pre-training supervision.} Success rate (\%) with
    post-training fixed at video\,+\,action and pre-training supervising
    nothing (\emph{None}), the video only, the actions only, or both, on (a)
    MimicDroid in-context, averaged over the four tasks per level with
    \textit{Avg.} over L1--L3, and (b) TableTop-Sim context-free.}
    \label{fig:ablation}
\end{figure}

\noindent\textbf{Ablation on pre-training modalities.}
We evaluate the modality variants in \S\ref{sec:anal_mod}, fixing post-training to joint action--video prediction (Fig.~\ref{fig:ablation} and \S\ref{app:ablation-full}).
Action-only pre-training outperforms video-only on in-context MimicDroid, particularly on L2 and L3, while the reverse holds on context-free TableTop-Sim.
MimicDroid preserves the pre-training action space and reference conditioning, whereas TableTop-Sim uses a different action space and no reference input.
Joint pre-training achieves the highest average success in both settings, with a small advantage over video-only on TableTop-Sim ($65.4$ versus $64.6$).
This pattern suggests that action supervision provides explicit guidance on the motion and temporal progression of interactions, yielding a prior that is readily reused when the action representation and conditioning are preserved.
Video prediction complements this signal by supervising the visual consequences of motion, potentially supporting transfer across larger differences from the pre-training setup.


\begin{figure}[t]
    \centering
    \includegraphics[width=\linewidth]{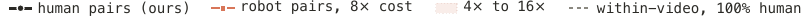}\\[0pt]
    \begin{subfigure}[t]{0.596\textwidth}
        \includegraphics[width=\linewidth]{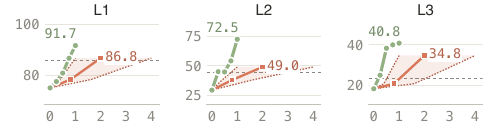}
        \caption{In-context}
    \end{subfigure}\hfill
    \begin{subfigure}[t]{0.396\textwidth}
        \includegraphics[width=\linewidth]{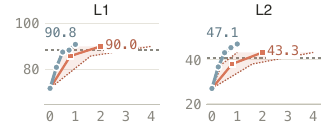}
        \caption{Context-free}
    \end{subfigure}
    \caption{\textbf{Scaling the pre-training data.} MimicDroid success rate (\%) against \rev{the collection budget (units of the full EgoVerse pair corpus) on a fixed Cosmos Policy backbone. Human pairs: $0$ to $100\%$; robot pairs: RoboMIND, at $8\times$ (line) or $4\times$ to $16\times$ (band) the cost of human video.} Dashed: within-video, all EgoVerse pairs.}
    \label{fig:scale}
\end{figure}

\noindent\textbf{Scaling the pre-training data.}
\label{sec:scaling}
We vary the amount of training data and its matched references while keeping the backbone and post-training setup fixed.
Larger datasets also use more pre-training steps (\S\ref{app:ladder}).
\rev{To weigh this against robot data, we also pre-train the same objective on RoboMIND robot pairs~\citep{wu2025robomind} and plot both on a collection-budget axis, as in RoboVQA~\citep{sermanet2024robovqa}.
Human demonstrations are collected about $10\times$ faster than robot teleoperation in EgoMimic~\mbox{\citep{kareer2025egomimic}} and $4.6\times$ faster in DexWild~\citep{tao2025dexwild}, and even a handheld gripper is over $3\times$ faster~\citep{chi2024universal}; we take $8\times$ as the central ratio and show $4\times$ to $16\times$.}
Performance improves with more data in both settings (Fig.~\ref{fig:scale}).
With $25\%$ of the cross-video data, \method matches within-video pre-training using all the data on the unseen in-context levels (L2 and L3).
It surpasses this baseline at $50\%$ on L3 and $75\%$ on L2.
Under context-free post-training, it exceeds the baseline on L2 at around $50\%$.
\rev{At $8\times$, human pairs reach higher success than robot pairs at the same or a lower budget on every level; on in-context L2, the full human corpus reaches $72.5\%$ and the largest robot setting $49.0\%$ at about twice the budget (\S\ref{app:robomind-post}).
Unlike Vid2Robot~\citep{jain2024vid2robot}, which needs a paired robot demonstration for every human one, \method pre-trains on human pairs alone, so it scales with human video~\citep{hoque2026egodex,punamiya2026egoverse}.}

\subsection{\rev{Real-Robot Evaluation}}
\label{sec:real}

\begin{figure}[t]
\centering
\begin{subfigure}[t]{\linewidth}
    \includegraphics[width=\linewidth]{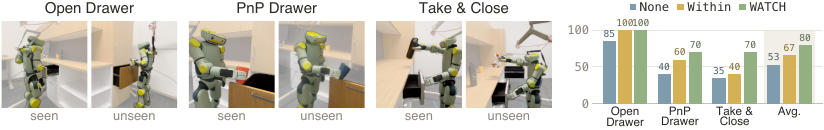}
    \caption{Context-free: \method in the seen (Easy) and unseen (Hard) setting of each task; success is their mean (Table~\ref{tab:real})}
    \label{fig:real-a}
\end{subfigure}\\[4pt]
\begin{subfigure}[t]{\linewidth}
    \includegraphics[width=\linewidth]{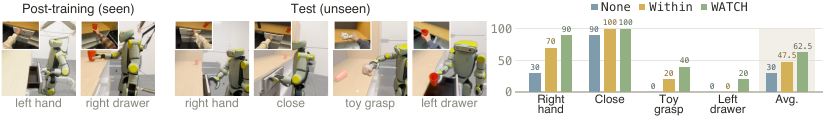}
    \caption{In-context: success on the unseen tasks, human video inset (Table~\ref{tab:real-icl})}
    \label{fig:real-b}
\end{subfigure}\\[4pt]
\begin{subfigure}[t]{\linewidth}
    \includegraphics[width=\linewidth]{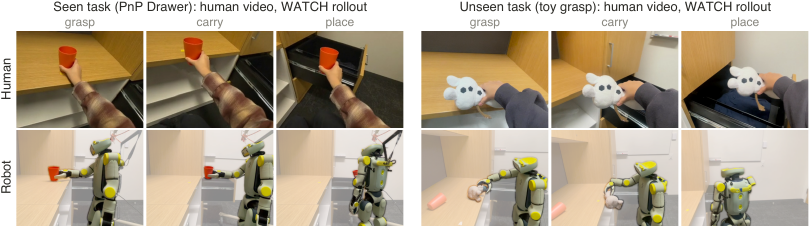}
    \caption{Rollout snapshots: human video (top) and \method (bottom), synced by task phase}
    \label{fig:real-c}
\end{subfigure}
\vspace{-0.6em}
\caption{\textbf{Real humanoid.} Success rate (\%, $10$ trials per setting). Avg.\ in (b) is the mean over the unseen tasks. In (c), \method rollouts on the seen PnP Drawer task (left) and on the unseen toy grasp (right), each below a human video of the same task.}
\label{fig:real}
\vspace{-0.8em}
\end{figure}

\rev{We evaluate on a physical humanoid~\citep{robotics2026fauna} performing whole-body manipulation, with two arms, parallel grippers, and an egocentric stereo camera; every checkpoint is post-trained on the same teleoperated demonstrations (App.~\ref{app:real}).
Context-free (Fig.~\ref{fig:real}a), with Hard settings that replace the drawer, the cup, or the tumbler, \method matches or exceeds within-video pre-training in every setting, which in turn improves over none; on Take \& Close Hard, within-video pre-training falls to the level of none ($10\%$) while \method reaches $50\%$.
In context (Fig.~\ref{fig:real}b), robot demonstrations of two tasks are paired with human videos, and at test time the policy must follow a human video of an unseen variant or task.
All models solve the seen tasks and closing the drawer, but \method best follows the other hand ($90\%$, against $70\%$ for within-video pre-training and $30\%$ for none) and the new toy grasp ($40$, $20$, and $0\%$; Fig.~\ref{fig:real}c), and it is the only one to place the cup in the left drawer ($20\%$).
With $10$ trials per setting, these results serve as a consistency check that the ordering from simulation carries over to hardware: the prior learned from human video alone transfers to a robot whose body, gripper, and camera all differ from the human demonstrators.}

\section{Related Work}
\label{sec:related}
\label{app:related}
\noindent\textbf{Pre-training for robot policies.}
Pre-training on large-scale, cross-embodiment robot datasets enables generalist policies to acquire diverse skills and transfer across embodiments \citep{kim2024openvla,nvidia2025gr00t,pi05}, but scaling this experience remains constrained by robot data collection.
Human videos offer additional experience through inferred latent actions \citep{ye2025lapa, garrido2026learning} or explicit motion supervision \citep{punamiya2026egoverse, kareer2025egomimic}.
Recent approaches co-train on human and robot data to ground this experience in executable actions \citep{punamiya2026egobridge, kareer2025emergence}.
Video-based models also incorporate action learning before target-robot adaptation: Cosmos3 introduces dynamics and policy mid-training \citep{nvidia2026cosmos3}, while ZimaBlue aligns video and action on heterogeneous robot trajectories after robot and human video pre-training \citep{wu2026zimablue}.
These co-training and alignment stages require robot data before target-domain adaptation.
More recent efforts explore the capabilities of large-scale human demonstrations and physical interaction data, including Dyna-2, GEN-1.5, and Skild S1 \citep{dyna2026dyna2,generalist2026gen15,skild2026s1}. 
They study human-video pre-training for world-action models and further report that their pretrained policies can adapt to new tasks from demonstrations provided in context, highlighting the potential of pre-training to support in-context robot learning.
\method instead uses task-matched human demonstrations alone during its proposed pre-training stage, followed by post-training on a small amount of target-domain data.

\noindent\textbf{Demonstration-conditioned robot learning.}
One- and few-shot imitation learning enables task adaptation from a small number of robot demonstrations \citep{fu2024context} or human videos
\citep{kim2025uniskill,jain2024vid2robot,xie2025human2robot,shah2025mimicdroid,bahety2024screwmimic,papagiannis2025rx, park2026retrieve}.
Most closely related, Zero-WAM learns from automatically constructed human-to-robot pairs to execute unseen tasks conditioned on human video prompts \citep{zhou2026zero}.
These approaches establish demonstrations as an effective interface for task adaptation; \method explores their role in representation learning through human-to-human pre-training.
The learned representations support both in-context control and reference-free behavior cloning after robot post-training.
This perspective draws on reference-conditioned learning in language modeling, image synthesis, and time-series forecasting \citep{borgeaud2022retro,blattmann2022rdm,liu2024ratd},
and examines its implications for transferable representations in robot foundation models.
\rev{Table~\ref{tab:related} in App.~\ref{app:related-table} places \method among these methods: demonstration-conditioned policies train across demonstrations but need a demonstration at test time, within-video world-action models predict video and actions without one but never across demonstrations, and \method does both from human pairs alone. UniSkill learns skill representations from unlabeled RGB video, without task labels or 3D hand trajectories, whereas \method relies on task labels for pairing and on hand motion, recorded or estimated from RGB (App.~\ref{app:label-noise}).}

\section{Conclusion}

We presented \method, a cross-video pre-training stage that turns task-paired
human demonstrations into a prior over motion and interaction for world-action
models, without matched human--robot data. Because the model must predict one
demonstration from another, it is trained toward how a task is executed rather
than how a demonstration looks: the learned features align the same motion across people,
rigs, and robots, attend to the hand and the object, and place robot frames where a model trained on robot data places them. This
prior carries over to control: across two backbones and three benchmarks it
improves both in-context and context-free post-training, most where the scene
departs from the demonstration, and it grows with paired \rev{egocentric} human video, which is
far cheaper to collect than robot data; extending the pairing \rev{to uncurated
video without task labels} is future work.

\subsection*{AI use statement}
In this work, we used generative AI tools to aid and polish writing, to draft
sections of the paper, for retrieval and discovery of related work, for
research execution, and for checking mathematical claims.
Drafted passages were rewritten by the authors and checked against the
experimental records; retrieved references were read and verified before being
cited; AI-assisted analysis and figure code was inspected, and its outputs were
cross-checked against the logged results by at least one author; and
derivations and statistical tests, including the pairing distance of
\S\ref{app:pairs} and the tests of \S\ref{sec:what-learns}, were
re-verified by hand. We have reviewed all AI-assisted work and take
responsibility for the final content of this work, including text, claims, and
artifacts produced with the aid of generative AI.

\subsection*{Acknowledgement}
We thank the NAVER AI Lab team for valuable discussions.
This work was done while Jeongeun Park was at NAVER AI Lab.
We also thank Fauna Robotics for their technical support throughout this work.

{\small
\bibliographystyle{unsrt}
\bibliography{references}}

\clearpage
\appendix

\section*{Appendix}
The appendix is organized as follows.
\begin{itemize}\setlength{\itemsep}{1pt}
    \item \S\ref{app:related-table} compares \method with related methods; \S\ref{app:limitations} states the limitations of this study.
    \item \S\ref{app:data} describes the data: the action and state representation (\S\ref{app:representation}), pair construction (\S\ref{app:pairs}), pairing statistics (\S\ref{app:stats}), the held-out sets (\S\ref{app:egoverse}), the MimicDroid robot data and evaluation protocol (\S\ref{app:mimicdroid}), and pair examples (\S\ref{app:pair-examples}).
    \item \S\ref{app:impl} gives implementation details: the architecture (\S\ref{app:architecture}), training and inference settings (\S\ref{app:training}), and the pre-training variants used as baselines (\S\ref{app:variants}).
    \item \S\ref{app:analyses} extends the analyses of \S\ref{sec:what-learns}: more frames for the motivation (\S\ref{app:motivation}), prediction on more held-out data (\S\ref{app:heldout-more}), the swapped-reference attention on more queries (\S\ref{app:cswap}), attention on robot and human frames (\S\ref{app:frames}), the protocol of the representation analyses (\S\ref{app:repr}), and additional probing results (\S\ref{app:probe-more}).
    \item \S\ref{app:robomind-mid} compares human video with robot data as pre-training data: the setup (\S\ref{app:robomind-setup}), the representation and attention analyses (\S\ref{app:robomind-analysis}), and downstream transfer from the two arms (\S\ref{app:robomind-post}).
    \item \S\ref{app:results} collects the baseline recipes (\S\ref{app:baselines}), per-task results (\S\ref{app:per-task}), the data scaling setup (\S\ref{app:ladder}), and the supervision ablation with the post-training target varied as well (\S\ref{app:ablation-full}).
    \item \S\ref{app:rollouts} shows qualitative rollouts.
\end{itemize}

\section{Comparison with Related Methods}
\label{app:related-table}
Table~\ref{tab:related} compares \method with demonstration-conditioned policies and within-video world-action models along the training objective (\S\ref{sec:related}).

\begin{table}[H]
\centering
\caption{\textbf{Cross-demonstration and joint video--action methods.}
Columns refer to the training objective. --: not applicable (no training
pairs); (\cmark): a stored demonstration is retrieved in place of a given one.}
\label{tab:related}
\small
\setlength{\tabcolsep}{5pt}
\renewcommand{\arraystretch}{1.12}
\newcommand{\grprel}[1]{\multicolumn{6}{l}{\rule{0pt}{8pt}{\footnotesize\textit{#1}}}\\[-1pt]}
\begin{tabular}{l|ccccc}
    \toprule
    Method & \makecell{Cross-demonstration\\training} & \makecell{Predicts\\video} & \makecell{Predicts\\actions} & \makecell{Pairs without\\robot data} & \makecell{Runs without\\test-time demo} \\
    \midrule
    \grprel{Demonstration-conditioned policies}
    Vid2Robot~\citep{jain2024vid2robot}       & \cmark & \xmark & \cmark & \xmark & \xmark \\
    UniSkill~\citep{kim2025uniskill}          & \xmark & \cmark & \xmark & --     & \xmark \\
    MimicDroid~\citep{shah2025mimicdroid}     & \cmark & \xmark & \cmark & \cmark & \xmark \\
    Human2Robot~\citep{xie2025human2robot}    & \cmark & \cmark & \cmark & \xmark & (\cmark) \\
    Zero-WAM~\citep{zhou2026zero}             & \cmark & \cmark & \cmark & \xmark & \xmark \\
    \grprel{Within-video world-action models and co-training}
    UWM~\citep{zhu2025uwm}                    & \xmark & \cmark & \cmark & --     & \cmark \\
    UVA~\citep{li2025unified}                 & \xmark & \cmark & \cmark & --     & \cmark \\
    Cosmos Policy~\citep{kim2026cosmospolicy} & \xmark & \cmark & \cmark & --     & \cmark \\
    EgoMimic~\citep{kareer2025egomimic}       & \xmark & \xmark & \cmark & --     & \cmark \\
    \midrule
    \rowcolor{ngreenlight}
    \method (Ours)                            & \cmark & \cmark & \cmark & \cmark & \cmark \\
    \bottomrule
\end{tabular}
\end{table}

\section{Limitations}
\label{app:limitations}

\paragraph{A shared Cartesian action space.} Human and robot actions are
written in one Cartesian task space, the end-effector pose and fingertip
positions of \S\ref{app:representation}, expressed in a canonical
frame fixed at the start of each episode. Three parts of the method rely on
this choice. The pairing distance (\S\ref{app:pairs}) compares wrist
trajectories and hand terms across demonstrations, which is meaningful only
because a human wrist and a robot wrist are described by the same
quantities. The reference slots condition the model on the reference's
actions in that space, so the action prior is a prior over end-effector
motion. And post-training reuses the same action representation, which is
why a human-trained action head can be warm-started on a robot at all. A
joint-space action target breaks all three at once: a human demonstration has no
counterpart to a robot's joints, so neither the pairing nor the action prior
carries over, and the prior would have to be re-expressed, through a
Cartesian intermediate target that a low-level controller tracks, or through
a learned map between joint and end-effector motion. All three post-training
targets in this paper, including the $20$-d bimanual ALOHA space, are
Cartesian, so the method covers end-effector controllers as they are commonly
deployed but has not been shown for joint-space policies. The canonical
frame carries a second assumption: the frame is anchored once per episode, so
a mobile base or a moving camera would need the frame to be re-anchored or
the motion to be expressed relative to the current view.

\paragraph{Pairing from a task taxonomy and tracked hands.} Candidates are
gated by a task class and ranked by a descriptor built from the hand
representation (\S\ref{app:pairs}), so pairing needs task labels and
hand motion. The motion need not come from tracking hardware: pre-training
on hands estimated from RGB alone keeps most of the gain
(Table~\ref{tab:label-noise-downstream}). The gain is smaller than with
tracked hands on in-context L2 and L3, where the policy must follow the
reference into a new scene, which suggests that the quality of the estimated
motion may affect reference-guided adaptation to new scenes. Unlabeled video, moreover, still
needs a label-free gate before it can be paired. The gate also
fixes the granularity of a pair: a class that mixes distinct motions admits
pairs that agree in task but not in motion, and the ranking has to separate
them, which it does well beyond same-class random pairs
(\S\ref{app:stats}) but not perfectly. The ranking compares motion
shape rather than magnitude, so a reference tells the model what to do but
not how far to move, and the extent has to come from the query's own frame
and state.

\paragraph{Room to scale.} Pre-training uses paired windows from a single
human dataset, and the scaling curve (\S\ref{sec:scaling}) has not
saturated: a quarter of the corpus already matches within-video
pre-training on the full corpus on the unseen levels, and every level is
still rising at the full corpus. More paired human video from more
collection sources is therefore the first lever, and the pairing places no
limit on how many sources it can span. Robot data is the second: cross-video
prediction applies to robot pairs as it does to human pairs
(Fig.~\ref{fig:tsne-pooled}), so adding robot demonstrations where they exist
should add to the prior rather than replace it; whether robot data helps
more when it enters through the same cross-video objective than alongside it,
as in co-training, is a direct comparison we leave to future work.

\section{Dataset Details}
\label{app:data}

\subsection{Action and State Representation}
\label{app:representation}

\paragraph{Layout.} The $57$-dimensional vector is laid out as
\[
\underbrace{\text{head}_{9}}_{0:9}\;\big|\;
\underbrace{\text{eef}^{L}_{9}}_{9:18}\;\big|\;
\underbrace{\text{hand}^{L}_{15}}_{18:33}\;\big|\;
\underbrace{\text{eef}^{R}_{9}}_{33:42}\;\big|\;
\underbrace{\text{hand}^{R}_{15}}_{42:57}.
\]
A pose term is $[\,p,\; R_{:,1},\; R_{:,2}\,]$, a translation followed by the
first two columns of the rotation matrix. A hand term is five fingertip
positions expressed in the wrist frame, in meters, which describes a grasp
without committing to a hand model. For a parallel gripper the same term is
synthesized by interpolating between a fully open and a fully closed fingertip
template with the normalized gripper command, so a gripper and a hand occupy the
same coordinates.

\paragraph{Frames.} For human demonstrations the canonical frame is placed at the
waist at the first frame of the episode, with its heading taken from the mean
head heading over the opening frames and its vertical axis aligned to gravity,
and is held fixed for the whole episode. Expressing the absolute form in the
moving head frame would instead make a stationary object appear to move whenever
the wearer turns. For robot demonstrations the pose terms are base-relative and the
head term is constant, since the cameras do not move with the arms. Each
source's channels are normalized by robust per-channel statistics estimated
once over its own windows.

\paragraph{\rev{Pose preprocessing.}} \rev{Human poses are used as released by EgoVerse, whose conversion
pipeline cleans the raw tracking in three steps before any window is formed.
Frames that jump more than $0.3$\,m from both neighbors, and missing or
placeholder frames, are replaced by linear interpolation. Each pose channel is
then filtered by a median filter followed by a Savitzky--Golay filter: a
$7$-frame median and a $15$-frame cubic filter for Aria, and a $5$-frame median
and a $9$-frame quadratic filter for MECKA and Scale. Finally, an episode is
discarded when more than $1\%$ of its frames place a hand over $2$\,m from the
head, which indicates misaligned SLAM origins; no episode in our corpus
exceeds this threshold.}

\subsection{Pair Construction}
\label{app:pairs}

\paragraph{Distance.} Within a task-taxonomy group, candidates are ranked by
\begin{equation}
d(q,r) = \min_{\pi \in \Pi_b} \frac{1}{|\pi|} \sum_{(i,j) \in \pi}
\left\| \frac{\phi_i(q) - \mu(q)}{\sigma(q)} - \frac{\phi_j(r) - \mu(r)}{\sigma(r)} \right\|_{W}
\;+\; \lambda_{h}\,\max\!\left(0,\, \left| h(q) - h(r) \right| - \tau\right) ,
\label{eq:pair}
\end{equation}
where $\phi_{1:T_d}(w)$ is the per-frame motion descriptor of window $w$ and
$\mu(w)$ and $\sigma(w)$ are its channel-wise mean and standard deviation
over the window's $T_d$ frames, the division taken per channel. This
standardization removes the window's offset and reduces the sensitivity of
the comparison to motion magnitude, so that executions of different reach are
compared by the shape of their motion; it is not a strict invariance, since
the position, rotation and hand channels do not all scale with reach. The
first term warps the two motions against each other within a band
$|i-j| \le b$, and $h(w)$, the left hand's share of the total reach,
separates a left-handed execution from a right-handed one, with a hinge at
$\tau$ so that small differences in hand use go unpenalized.

\paragraph{Descriptor.} Per frame and per hand, the descriptor concatenates the
start-relative position ($3$), the $6$D rotation ($6$), the action increment of
the end effector ($9$) and the hand term ($15$), giving $33$ channels per hand
and $66$ per frame. The sequence is subsampled to $T_d = 40$ steps before
matching, and the weights $W$ of Eq.~\ref{eq:pair} down-weight the fingertip
channels relative to the wrist so that the trajectory rather than the finger
detail drives the alignment: the fingertip channels carry a weight of $0.25$
relative to the wrist. The band is $b = 6$ steps, and the handedness term uses
$\lambda_{h} = 3.0$ with threshold $\tau = 0.25$; it separates a left-handed
from a right-handed execution of the same task, and path length is not used for
this, since it does not distinguish the two.

\paragraph{Reference set.} Candidates are first gated to the query's task class
in a canonical taxonomy, falling back to the class's super-group only when no
other source recorded that class. Within the gate, a query keeps its $8$ nearest
references from the same collection source and up to $7$ from other sources, a
cross-source candidate being admitted whenever its distance is within $1.6$
times the best score, for at most $15$ references per query (the manifest
reserves $20$ slots). The full
build gives $1{,}092{,}145$ windows and $1{,}045{,}007$ query windows with an
average of $10.5$ references each (median $11$, $4$ to $15$ between the 5th and
95th percentiles), $4.2$ from the same source and $6.4$ from others; no query
is without a reference, and $9.3\%$ have no cross-source one: queries the
super-group fallback does not reach, either because no other source recorded
their super-group or because no cross-source candidate fell inside the
distance gate. They are all MECKA queries, whose bridging rate is $82.4\%$
against $100\%$ for Scale, Aria and the play set. At each training step
one reference is drawn uniformly from the query's whole set, from any source,
so a query with cross-source references is conditioned on one of them in
proportion to their share of its set.

\subsection{Data Statistics}
\label{app:stats}

\begin{table}[h]
\centering
\caption{\textbf{Within a task class, the retrieved reference agrees with the query's motion.}
Agreement between a query and its top-ranked cross-source reference on
motion-agreement metrics that the pairing score does not directly optimize
(the descriptor sees positions, rotations and action increments, but is
scored by warped descriptor distance, not by these metrics), against a random
partner of the same task class and a random partner. Means over $12{,}000$
queries.}
\label{tab:pairing}
\small
\setlength{\tabcolsep}{7pt}
\renewcommand{\arraystretch}{1.1}
\begin{tabular}{l|ccc}
    \toprule
    & Retrieved & Same-class null & Random null \\
    \midrule
    speed-profile correlation $\uparrow$      & \textbf{0.23} & $-0.01$ & $-0.01$ \\
    trajectory-shape cosine $\uparrow$        & \textbf{0.36} & 0.00 & 0.00 \\
    $|\log|$ path-length ratio $\downarrow$   & \textbf{0.69} & 0.93 & 0.92 \\
    speed off by more than $3\times$ $\downarrow$ & \textbf{22.9\%} & 28.7\% & 39.4\% \\
    \bottomrule
\end{tabular}
\end{table}

\paragraph{Corpus.} Pre-training queries come from the three human collection sources of
EgoVerse, MECKA, Scale and Aria, all at $30$\,Hz in the $57$-d convention of
\S\ref{app:representation}, together with the MimicDroid play at
$20$\,Hz, which is human as well: hand-tracked play by people in the sink
environment, with no robot in the loop (\S\ref{app:mimicdroid}). The
manifest holds $14{,}137$ episodes, $13{,}315$ of which carry query windows
(the rest hold reference windows only), and $1{,}045{,}007$
query windows, of which $551{,}337$ are MECKA, $367{,}823$ Scale, $109{,}652$
Aria and $16{,}195$ play, each query carrying up to $20$ reference slots, so
every query is a human demonstration. The robot subset of EgoVerse and
Humanoid Everyday, both robot data, are listed as windows but serve as
neither query nor reference here; EgoDex is in no training manifest at all.

\paragraph{Windows and totals.} Pre-training windows span about $3$ seconds: $92$ frames
at $30$\,Hz for the three EgoVerse sources and $60$ frames at $20$\,Hz for
the $20$\,Hz sources, the MimicDroid play and the robot windows, the lengths
that pack with the current frame into the video autoencoder's $1 + 4(T-1)$
frames (\S\ref{app:architecture}) and that the action chunk of
\S\ref{app:training} follows. Windows are drawn at a stride of $30$
frames, with at most two windows per episode among a query's references, and
the query's own episode is excluded from its references except in the
MimicDroid play, whose twenty episodes are its only source; its $19{,}802$
links are the only same-episode links in the build.
The snapshot we pair holds $14{,}137$ episodes of the $1.36$K-hour EgoVerse
release, with the release's robot subset left out (it enters only under
co-training, \S\ref{app:variants}). At the
$1$-second stride the $1{,}045{,}007$ query windows amount to $871$ hours of
$3$-second windows; with the one reference drawn per query at each step, an
epoch covers $1{,}742$ hours, and the full set of $10.99$M query--reference
links $10.0$K hours. These are window-hours: each query window and each
reference link counts as one $3$-second window, and the hours of
\S\ref{app:robomind-mid} are counted the same way.

\paragraph{Pairing statistics.} The build holds $10.99$M query-reference
links. They cover the corpus rather than a corner of it: $90.6\%$ of the
episodes ($68.0\%$ of the windows) serve as a reference at least once, a
query's $20$ slots span $2.0$ sources and $10.4$ episodes on average, and the
reference stream the model actually sees, weighted by how often each window
is drawn, retains $95\%$ of the pool's motion diversity ($247.8$ against
$260.5$ effective motions under an RBF Vendi score on the descriptor). The
DTW rerank adds motion fidelity on top of the taxonomy gate, which
Table~\ref{tab:pairing} scores on additional motion-agreement metrics that
the pairing score does not directly optimize,
against a null that draws a random partner of the same class. There are no
self-matches, no overlapping windows between a query and its references, and
no episode shared between training queries and the held-out set.

\subsection{Held-Out Data}
\label{app:egoverse}


\paragraph{Split.} The held-out set is $150$ episodes, stratified over task and
source and taken at the level of whole episodes, so no window of an
evaluation episode is used as a training query.

\paragraph{Held-out sets.} Two sets are kept out of training. (i) $150$
whole episodes of the training sources ($75$ MECKA, $67$ Scale, $8$ Aria),
stratified over task and source, are removed from the manifest and become the
held-out EgoVerse queries ($13{,}369$ windows) behind the analyses of
\S\ref{sec:what-learns}; their references are retrieved from the
training pool. (ii) EgoDex~\citep{hoque2026egodex}, recorded on a different
rig with tracked fingers and converted to the same $57$-d convention at
$30$\,Hz, is the held-out human source of Table~\ref{tab:offline-more}: it
enters no training manifest, is paired only against itself by window-level
top-$3$ retrieval, and carries its own normalization. Table~\ref{tab:offline}
and Table~\ref{tab:offline-more} use $2{,}048$ clips per set.

\paragraph{Analysis sets.} The analyses of \S\ref{sec:what-learns} are
drawn from the $150$ held-out episodes. The conditioning test
(Fig.~\ref{fig:robustness}) needs, for each query, a matched demonstration
from a different source and a different-task demonstration from the same source, and
keeps the $137$ episodes for which both exist.
The representation analyses (Fig.~\ref{fig:invariance}) need pairs of queries
that perform the same motion, and define motion similarity through the training
retrieval itself: two held-out queries are a pair when their top-$20$ references
share a demonstration, which is the retrieval descriptor's own notion of the
same motion. Each query is then read with its own references, so neither member
of a pair enters the other's input and nothing leaks. Restricting to tasks that
at least two sources recorded, since a task recorded by a single source confounds
task with identity, leaves $122$ episodes, $359$ action windows and $120$
cross-source pairs, on which the human-set retrieval and probing results are
computed; the t-SNE of \S\ref{app:repr} draws $95$ of these windows,
from three of the tasks that two of the three sources recorded. The representation analyses are also run on a robot window set that is not
drawn from EgoVerse: $396$ RoboMIND windows from three robots and six tasks,
with windows of the same length and pairs matched by trajectory distance
within a task, described in \S\ref{app:repr} and illustrated in
Fig.~\ref{fig:robomind-example}.

\subsection{MimicDroid Robot Data}
\label{app:mimicdroid}

\paragraph{What the release provides.} MimicDroid ships the human side of the
benchmark: hand-tracked human play in the sink environment ($20$ episodes at
$20$\,Hz), which is the play in the
pre-training manifest of \S\ref{app:stats}, and three canonical
demonstrations per task, twelve in all. None of it is robot-executed: the
release ships no robot training episodes, so the GR1 training set, the
retrieval pool and the evaluation scenes below are ours.

\paragraph{From human demonstrations to robot episodes.} Every GR1 training
episode comes from one of those twelve human demonstrations, augmented in the
MimicGen manner and then retargeted. A scene is drawn by a seeded environment
reset in a (layout, style) cell; one of the task's three human demonstrations
is retargeted into that scene, and the retargeted trajectory is replayed on
the GR1 humanoid, the episode being kept only if the replay succeeds. So the
three demonstrations per task are the only human motions in the set, and the
scene draws are what multiply them into $253$ robot episodes. Retargeting
carries the demonstrated two-hand end-effector path into the drawn scene and
leaves the arm configuration to the controller: the replay commands absolute
base-frame poses of the two end effectors together with their gripper signals,
which an operational-space controller resolves on the humanoid. Each episode
is an open-loop approach to the demonstration's start at $1$\,cm per frame, a
hand-off blend, and then the demonstration replayed unaltered, with success
judged once at the end; approach and blend are $30$ and $10$ frames for the
door task, $20$ and $10$ for the plate task, and $0$ and $16$ for the cabinet
and faucet tasks.

\paragraph{Timing around the grasp.} The humanoid's gripper cannot close as
fast as the demonstrated hand and still hold the object, so the retargeted
trajectory is given more time there: for the two pick-and-place tasks it holds
for $12$ frames before the gripper closes and stretches the closing itself
over twice its recorded duration. Without this the plate task replayed
successfully in $1$ scene draw out of $6$; with it, in $5$ out of $6$.

\paragraph{Scenes and objects.} Objects spawn within $4$\,cm of their position
in the source demonstration and are restricted to one species per task (a
tomato for the cabinet task, corn for the plate task), since the recorded grip
is tied to the object's size. The same feasibility filter is applied to
training and evaluation scenes, so it defines the distribution rather than
biasing it. One scene draw records an episode of \emph{every} task feasible in
that layout, so the scene and the proprioception alone do not determine what
is to be done and only the demonstration does.

\paragraph{Sizes.} The set holds $253$ episodes over the four L1 tasks ($50$
\textsc{TurnOnFaucet}, $50$ \textsc{CloseLeftCabinetDoor}, $81$
\textsc{PnPSinkToCabinet}, $72$ \textsc{PnPSinkToRightCounterPlate}), each
$256$ to $302$ frames at $20$\,Hz with three cameras, and a retrieval pool of
$100$ demonstrations per task drawn from further scene draws. A second copy of
the pool replayed at double speed serves the infeasibility test of
\S\ref{sec:setup}, for post-training and evaluation alike; the query
side is unchanged, and the matching descriptor drops its velocity channels,
which the faster replay would otherwise double. Windows span $36$ frames at
$20$\,Hz, the $32$-step action chunk of \S\ref{app:training} plus four frames
of headroom so that the reader can round the span down to the video
autoencoder's $1 + 4(T-1)$ lengths (\S\ref{app:architecture}) as the
pre-training windows do (\S\ref{app:stats}); they are taken at every
time step, giving $61{,}686$ query windows.

\paragraph{Pairing.} Each query window carries six demonstration slots. Slot
$0$ is the episode's own canonical demonstration, aligned to the query by a
global DTW path over the whole episode, which is the conditioning the one-shot
evaluation supplies at every chunk; slots $1$ to $5$ are the five nearest pool
windows under banded DTW on a $26$-d two-hand pose descriptor of the preceding
$32$ frames, at most one per episode. Because robot and demonstration share
units and a frame, this matching runs on absolute poses and needs none of the
per-window normalization of Eq.~\ref{eq:pair}. Data, targets and normalization
are identical under in-context and context-free post-training, which only drop
the reference slots; the action targets of the two readers agree to the bit.

\paragraph{Evaluation.} L2 holds out the style and the objects within the
training layout (\textsc{PnPSinkToRightCounterPlate}, \textsc{PnPSinkToCabinet},
\textsc{CloseLeftCabinetDoor}, \textsc{CloseRightCabinetDoor}) and L3 holds out
the layout (\textsc{TurnOnFaucet}, \textsc{PnPSinkToRightCounterPlate},
\textsc{CloseLeftCabinetDoor}, \textsc{PnPSinkToMicrowaveTop}). Evaluation
scenes are frozen manifests whose seed stream is disjoint from training
(evaluation seeds start at $1{,}000$ and training seeds at $1{,}000{,}000$, and
the runtime refuses an overlap), and neither L2 nor L3 scenes are ever trained on; the context is one canonical
demonstration recorded in the held-out setting. Each task is evaluated over
$30$ episodes, $10$ per demonstration for each of its three canonical
demonstrations, except L2 \textsc{PnPSinkToRightCounterPlate}, which has four
demonstrations and $40$ episodes; the success rates of Tables~\ref{tab:icl-main}, \ref{tab:bc-main}
and \ref{tab:ablation} are means over these episodes. The policy acts in $32$-step
chunks, generated with $35$ denoising steps at guidance $1.5$. Under in-context post-training the demonstration window is re-retrieved
at every chunk from the policy's own recent proprioception, constrained to
advance monotonically, within a band of $8$ to $48$ frames past the previous
match, and locked to one demonstration episode, so that one
demonstration conditions the whole episode; context-free post-training
bypasses retrieval altogether.

\paragraph{Post-training target.} The robot arm predicts the commanded action in
the controller's frame rather than the measured state. Predicting the measured
state and replaying it as a command would apply the tracking lag twice, once in
the demonstration and once in the rollout, and the commanded form also needs no
frame conversion at execution time.

\subsection{Pair Examples}
\label{app:pair-examples}

\begin{figure}[p]
    \centering
    \begin{subfigure}{\linewidth}
        \centering
        \includegraphics[width=\linewidth]{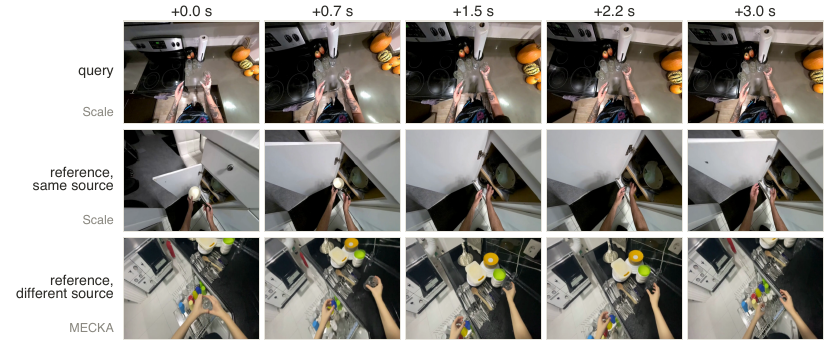}
        \caption{Unload dishwasher}
    \end{subfigure}\\[3pt]
    \begin{subfigure}{\linewidth}
        \centering
        \includegraphics[width=\linewidth]{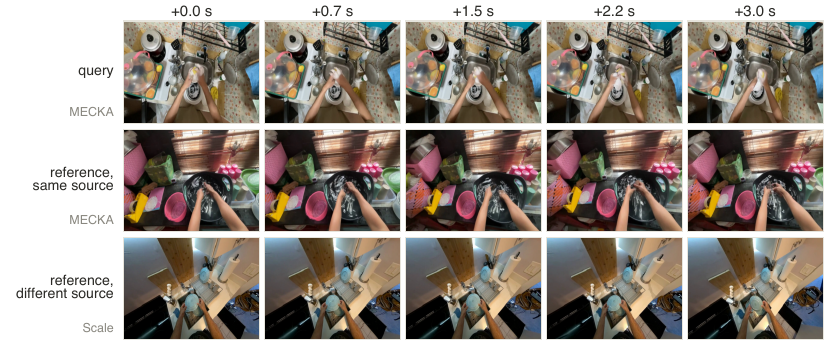}
        \caption{Do dishes}
    \end{subfigure}\\[3pt]
    \begin{subfigure}{\linewidth}
        \centering
        \includegraphics[width=\linewidth]{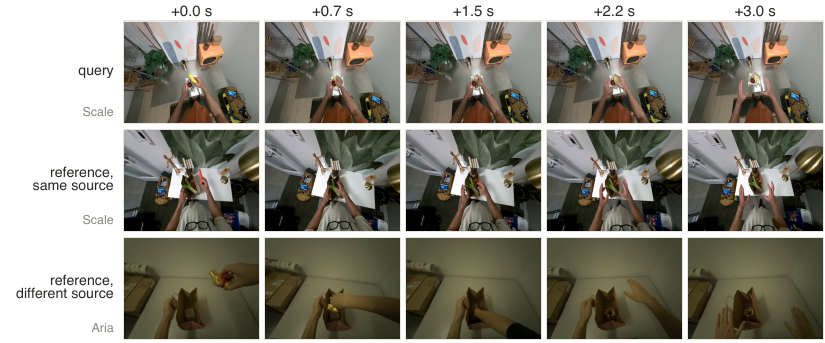}
        \caption{Bagging groceries}
    \end{subfigure}
    \caption{\textbf{Query--reference pairs.} Three queries from the training
    pairs, each with its reference from the same collection source and one from
    a different source, as five frames spanning the $3$-second window. The source of
    every clip is named under its row label. (a) Unload dishwasher, a Scale
    query with a MECKA reference; (b) do dishes, a MECKA query with a Scale
    reference; (c) bagging groceries, a Scale query with an Aria reference.}
    \label{fig:pair-examples}
\end{figure}

Fig.~\ref{fig:pair-examples} shows three query--reference pairs as the
pairing produces them. Each query is a $3$-second window; its reference set
holds the nearest windows of the same task from the query's own collection
source and, by design, from different sources (\S\ref{app:pairs}). The
same-source reference shares the query's rig, viewpoint and environment; the
different-source reference shares only the task and the motion, with a different
person, camera and scene, which is the variation the cross-video objective
trains against.

\begin{figure}[t]
    \centering
    \includegraphics[width=\linewidth]{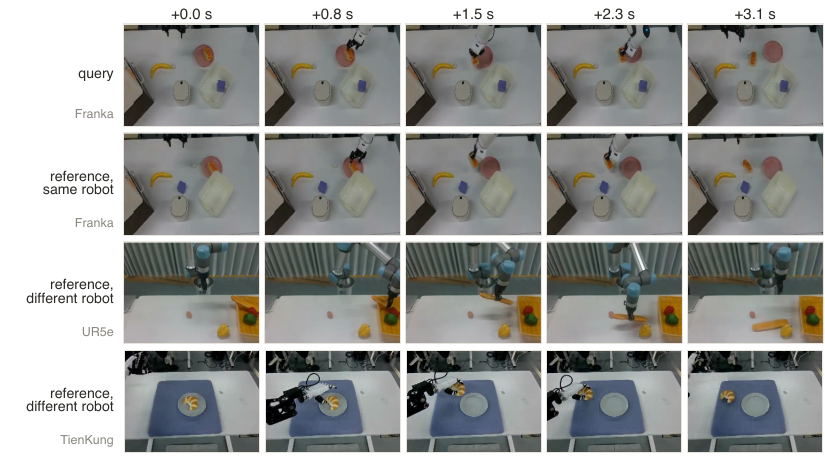}
    \caption{\textbf{A query of the robot analysis set with its nearest
    matches.} Bread in basket on RoboMIND: a Franka query window and the nearest
    same-task window, by dynamic-time-warping distance on the wrist trajectory,
    from the same robot and from each of the two other robots, as five frames
    over the $3$-second window. Cross-robot pairs like the last two rows are the
    queries of the robot retrieval test (Fig.~\ref{fig:invariance}b).}
    \label{fig:robomind-example}
\end{figure}

\section{Implementation Details}
\label{app:impl}

\subsection{Architecture}
\label{app:architecture}

\paragraph{Slot layout.} Pre-training packs six slots into the sequence, with
$k$ reference video slots and $g$ future video slots:
\begin{center}
\small
\begin{tabular}{llll}
    \toprule
    Slot & Contents & Time & Status \\
    \midrule
    0 & query state $s^q_t$ & $t$ & condition \\
    1 & reference actions $a^r_{u:u+C}$ & $t$ & condition \\
    2 & reference video $I^r_{u:u+C+1}$ & $t-k\,\delta \ldots t-\delta$ & condition \\
    3 & query frame $I^q_t$ & $t$ & condition \\
    4 & query actions $a^q_{t:t+C}$ & $t$ & generated \\
    5 & query future video $I^q_{t+1:t+C+1}$ & $t+\delta \ldots t+g\,\delta$ & generated \\
    \bottomrule
\end{tabular}
\end{center}
Query indices run over $t$ and reference indices over $u$, the start of the
reference window, as in Eq.~\ref{eq:cross}; the Time column is the
positional-time tag of each slot, which places the reference video before the
query frame regardless of $u$. Pixels are packed as $1 + 4(T-1)$ frames to respect the causal video
autoencoder, with black frames written into the non-pixel slots and overwritten
in latent space by tiling. The query frame is placed immediately before the
generation boundary so that the backbone's video continuation prior is preserved.

\paragraph{Backbone modifications.} Two changes are made to the video backbone.
Positional encoding is made continuous in time, taking a per-slot timestamp in
seconds instead of an integer frame index, so that a slot's position no longer
shifts when the number of reference or future frames changes. A slot-type embedding
is added to each slot from a small table, initialized at zero so that the
modified network reproduces the original one exactly at initialization; the
attention is bidirectional, so a condition placed after the generated slots is
still visible to them. Variable sequence lengths, which arise because human and
robot windows differ in frame count, are handled by grouping samples of equal
shape into a batch.

\paragraph{Instantiation on each backbone.} On the rectified-flow diffusion
transformer, slots are latent frames and a condition is supplied as a clean
latent excluded from the loss. On the mixture-of-transformers model there is no
separate conditioning encoder: every modality is packed into one token sequence
whose relations are carried by a shared 3D positional grid, the reference and
the query are placed on the same temporal grid so that their frames correspond,
and a condition is expressed by holding a token's noise level at zero. Actions
enter through a domain-aware linear map in that case rather than by tiling.
The task description $\ell$ of Eq.~\ref{eq:cross} is supplied as text on both
backbones, through T5 cross-attention on Cosmos-Predict2.5 and as caption
tokens on Cosmos3; the context-free variant drops the reference clause from
the caption.

\subsection{Training}
\label{app:training}

\paragraph{Cosmos3 pre-training for RoboCasa.} For the RoboCasa experiments,
the Cosmos3 pre-training mixture adds two RoboMIND streams~\citep{wu2025robomind}, Franka ($80$ tasks, $11{,}480$
episodes, $1.61$M frames, $10$-d actions) and TienKung ($32$ tasks, $2{,}785$
episodes, $4.6$M frames, $20$-d actions), each with three RGB views, beside
the EgoVerse pairs ($32.4$M frames). The three sources are mixed by rank at $28{:}2{:}2$, so RoboMIND is
$12.5\%$ of the training samples. Robot windows use an action chunk of $16$
frames ($0.53$\,s at $30$\,Hz), served without a reference, as RoboCasa is
a context-free benchmark. The Cosmos3 runs for MimicDroid and TableTop-Sim
use the human pairs alone.

\paragraph{Cosmos Policy recipe.} Compared with the original Cosmos Policy
recipe, our variant updates the video backbone from Predict2 to Predict2.5 and
reduces the post-training objective to future-video and action prediction,
without auxiliary future-proprioception prediction or rollout-data
augmentation.

\paragraph{Hyperparameters.} Pre-training uses a learning rate of $10^{-4}$ with
a $500$-step warmup and a cosine schedule (a $50$K-step cycle with a floor
at $0.3$ of the peak) for $8$K iterations, a global batch of $64$ ($2$
per GPU on $32$ GPUs); this is the full-data arm of the scaling sweep of
\S\ref{app:ladder}, whose smaller arms scale the cycle with the data. The
action weight is $\lambda = 48$. Post-training runs $8$K iterations with the
cycle matched to that length. Every configuration is a single training run. The action chunk
is $92$ frames at $30$\,Hz for human windows and $32$ frames at $20$\,Hz for the
MimicDroid and TableTop-Sim robot windows, which is why those windows span
$36$ frames (\S\ref{app:mimicdroid}); the RoboCasa robot chunk is the $16$
frames given above. On
Cosmos-Predict2.5 the full transformer is updated; on Cosmos3 the generation
experts, the time embedder, the latent and action projections, and the action
modality embedding are updated at a learning rate of $2 \times 10^{-4}$, with a
$5\times$ multiplier on the action heads, and the base model's action head is
loaded rather than re-initialized.

\subsection{Pre-Training Variants}
\label{app:variants}

The within-video baseline
keeps the architecture, the slot roles, the time axis and the objective of
\S\ref{sec:cross_video}, blanks the reference video and action slots and
shortens the reference video block to two slots, so the model is trained to
predict each window's actions and future frames from its own observation alone;
its sequence holds $29$ latents against $50$ for \method, so at the same
$8$K iterations it costs less per step. Co-training is the same blanked model with the two
robot sources promoted to queries, $33{,}176$ windows from the EgoVerse robot
subset and $92{,}136$ from Humanoid Everyday, for $1{,}170{,}319$ queries in
all; since the reference content is masked, adding a source needs no
pairing.

\section{Additional Analyses}
\label{app:analyses}

\subsection{More Frames for the Motivation}
\label{app:motivation}

\begin{figure}[!t]
    \centering
    \includegraphics[width=0.76\linewidth]{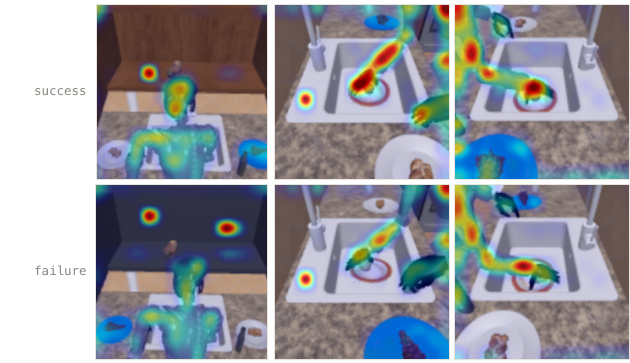}
    \caption{\textbf{The post-trained original video model, more frames.}
    Attention to the current frame of the two {\small\texttt{PnPSinkToCabinet}}
    rollouts of Fig.~\ref{fig:motivation}a, one frame later, from the three
    cameras. The success (top) attends to the hand at the object,
    the failure (bottom) to the forearm.}
    \label{fig:app-posttrained}
\end{figure}

Fig.~\ref{fig:motivation}a shows the original video model post-trained on
the robot at one frame of a successful and a failed rollout;
Fig.~\ref{fig:app-posttrained} gives the next frame of the same two rollouts
from all three cameras, with the same reading: the success attends to the hand at
the object, the failure to the forearm. Fig.~\ref{fig:motivation}b and c
show one human and one robot frame under the original video model and ours.
Fig.~\ref{fig:app-frames} (\S\ref{app:frames}) adds the within-video
baseline on four more robot frames and on three human queries. The reading
does not change with the frame: the backbone
spreads over edges and background, within-video pre-training moves the mass
onto the arm, and cross-video pre-training puts it on the hand and the object.

\subsection{Prediction on More Held-Out Data}
\label{app:heldout-more}

\begin{figure}[p]
    \centering
    \begin{subfigure}{\linewidth}
        \centering
        \includegraphics[width=0.94\linewidth]{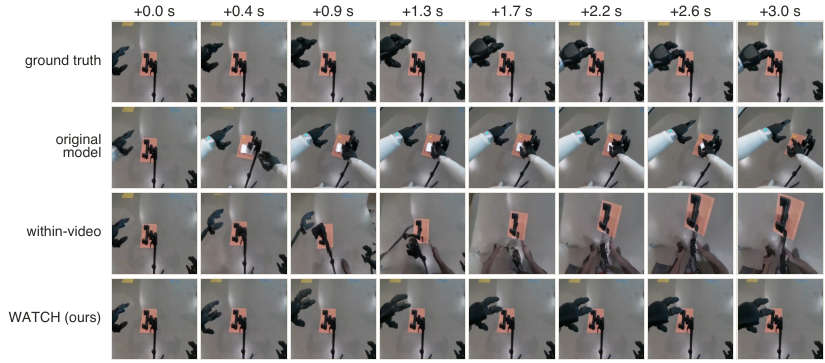}
        \caption{Humanoid Everyday}
    \end{subfigure}\\[3pt]
    \begin{subfigure}{\linewidth}
        \centering
        \includegraphics[width=0.94\linewidth]{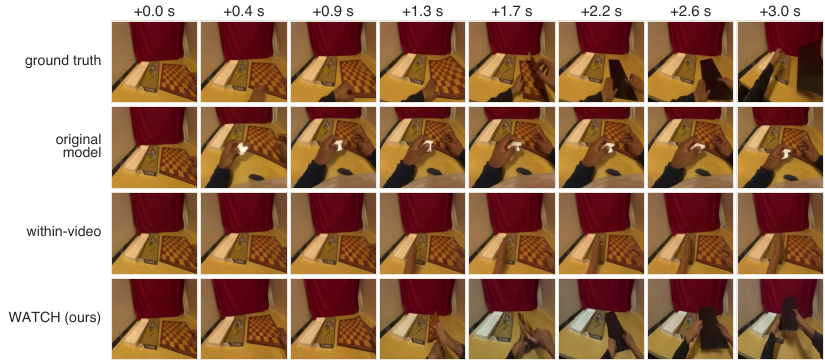}
        \caption{EgoDex}
    \end{subfigure}\\[3pt]
    \begin{subfigure}{\linewidth}
        \centering
        \includegraphics[width=0.94\linewidth]{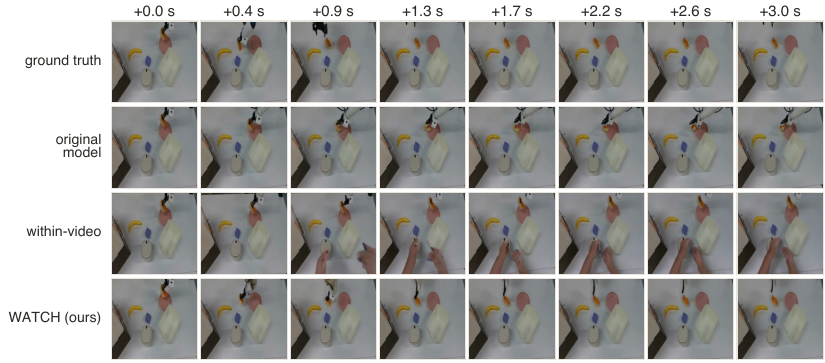}
        \caption{RoboMIND}
    \end{subfigure}
    \caption{\textbf{Generated video on the held-out sets.} One clip per set,
    (a) Humanoid Everyday, (b) EgoDex, (c) RoboMIND, as eight frames over the
    $3$-second generated window: the ground truth, the original video model,
    within-video pre-training and \method.}
    \label{fig:heldout-video}
\end{figure}

\begin{table}[H]
\centering
\footnotesize
\setlength{\tabcolsep}{4pt}
\renewcommand{\arraystretch}{1.08}
\caption{\textbf{Prediction on three held-out sets.} Table~\ref{tab:offline}
extended to Humanoid Everyday~\citep{zhao2025humanoid}, a humanoid
teleoperation dataset, and EgoDex, a second human set, $2{,}048$ clips per
set, with the co-training variant of \S\ref{app:variants} added. Action metrics are on the end-effector subset of the action: MSE in
normalization space and accuracy at tolerance $0.1$. The original model
generates video only. Best in bold, second underlined.}
\label{tab:offline-more}
\begin{tabular}{ll|cc|ccc}
    \toprule
    & & \multicolumn{2}{c|}{Action (end effector)} & \multicolumn{3}{c}{Generated video} \\
    \cmidrule(lr){3-4} \cmidrule(lr){5-7}
    Held-out set & Model & MSE $\downarrow$ & acc@$0.1$ $\uparrow$ & FVD $\downarrow$ & FID $\downarrow$ & LPIPS $\downarrow$ \\
    \midrule
    \multirow{4}{*}{Humanoid Everyday}
    & Original model & -- & -- & 325.7 & 31.81 & 0.422 \\
    & Within-video   & 0.077 & 51.5 & 610.4 & 31.22 & 0.366 \\
    & Co-training    & \underline{0.045} & \underline{66.2} & \textbf{105.1} & \underline{12.70} & \underline{0.267} \\
    \rowcolor{ngreenlight}
    \multicolumn{1}{l}{} & \method (Ours) & \textbf{0.023} & \textbf{71.4} & \underline{137.6} & \textbf{10.01} & \textbf{0.215} \\
    \midrule
    \multirow{4}{*}{EgoDex}
    & Original model & -- & -- & 200.5 & \textbf{8.18} & 0.243 \\
    & Within-video   & \underline{0.024} & \underline{69.0} & \underline{169.8} & 9.82 & \underline{0.242} \\
    & Co-training    & 0.033 & 65.8 & 220.1 & 14.17 & 0.299 \\
    \rowcolor{ngreenlight}
    \multicolumn{1}{l}{} & \method (Ours) & \textbf{0.019} & \textbf{73.4} & \textbf{142.7} & \underline{8.53} & \textbf{0.211} \\
    \midrule
    \multirow{4}{*}{RoboMIND}
    & Original model & -- & -- & \underline{290.7} & \underline{12.48} & \underline{0.247} \\
    & Within-video   & \underline{0.046} & 55.6 & 1485.1 & 53.53 & 0.370 \\
    & Co-training    & 0.056 & \underline{56.7} & 418.7 & 31.10 & 0.291 \\
    \rowcolor{ngreenlight}
    \multicolumn{1}{l}{} & \method (Ours) & \textbf{0.039} & \textbf{68.5} & \textbf{154.9} & \textbf{8.33} & \textbf{0.154} \\
    \bottomrule
\end{tabular}
\end{table}

Table~\ref{tab:offline} evaluates prediction on RoboMIND; Table~\ref{tab:offline-more}
adds two sets, Humanoid Everyday, a humanoid teleoperation dataset and thus
robot data, and EgoDex, human demonstrations from a different egocentric rig,
with the same end-effector action metrics and with the co-training variant
added. Co-training is not a held-out comparison on Humanoid Everyday:
it draws its robot queries from that dataset (\S\ref{app:variants}),
so it is tested on a source it was trained on, while \method, within-video
and the original model have never seen it. Even so, \method matches or
exceeds co-training on Humanoid Everyday on every metric but FVD, with half
its end-effector error, a higher accuracy and sharper frames by FID and
LPIPS: the cross-video objective reaches, on data it has
never seen, the level that co-training reaches by training on it. On EgoDex, which none of the models has seen,
\method leads on every action metric and on FVD and LPIPS, and co-training
falls behind within-video throughout. On RoboMIND, robot data that none of the models
has seen, \method again leads on every metric, with the lowest end-effector
error and an accuracy twelve points above the next model, and the sharpest
frames ahead of the original model, while within-video and co-training fall
well behind on video. Fig.~\ref{fig:heldout-video} shows one
clip from each set: on Humanoid Everyday \method tracks the ground-truth
gripper where the original model and within-video drift to a different arm
and scene, and on EgoDex it is the closest to the ground truth, lifting the
object as the hand does where the others leave it on the table.

\subsection{Swapped Reference on More Queries}
\label{app:cswap}

\begin{figure}[p]
    \centering
    \begin{subfigure}{\linewidth}
        \centering
        \includegraphics[width=\linewidth]{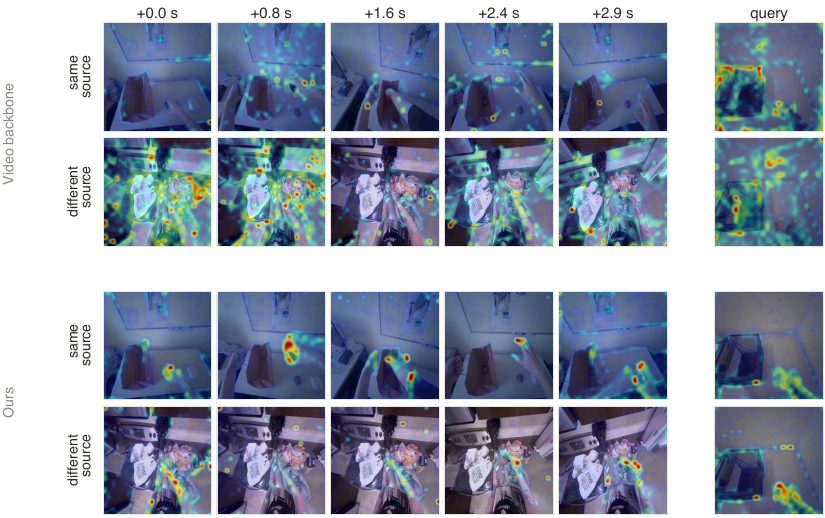}
        \caption{Bagging groceries}
    \end{subfigure}\\[3pt]
    \begin{subfigure}{\linewidth}
        \centering
        \includegraphics[width=\linewidth]{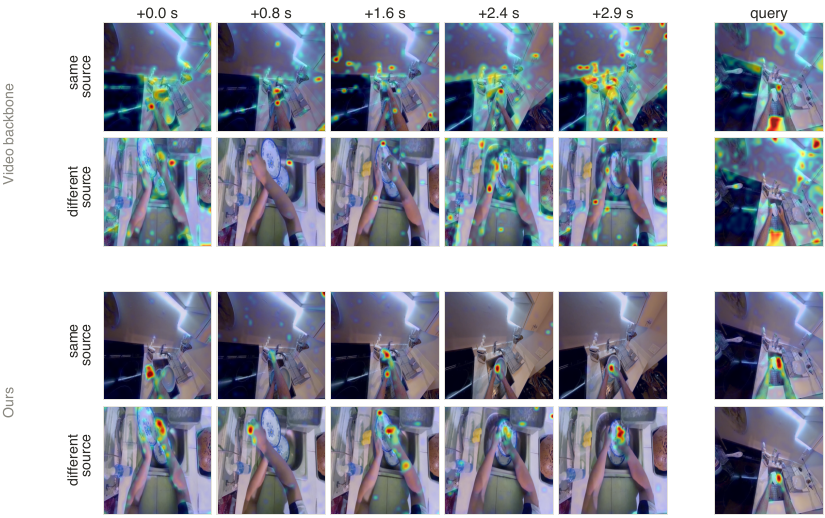}
        \caption{Do dishes}
    \end{subfigure}
    \caption{\textbf{Swapped reference, two queries.} Action-token attention
    of the video backbone (upper block of each query) and ours (lower block)
    with a reference from the same collection source and from a different one.
    (a) Bagging groceries: an Aria query, the different-source reference a
    Scale demonstration. (b) Do dishes, the query of Fig.~\ref{fig:robustness}b:
    a Scale query, the different-source reference a MECKA demonstration.}
    \label{fig:cswap}
\end{figure}

\paragraph{More queries and the backbone under a swapped reference.}
Fig.~\ref{fig:robustness}b shows the action-token attention of our model for
one query under a reference from the same collection source and from a different one.
Fig.~\ref{fig:cswap} repeats the comparison
on two more queries from different tasks and collection sources, and adds
the original video model under the same two references. The layout follows
Fig.~\ref{fig:robustness}b: the five columns on the left are the reference
at successive points along its span, the column on the right is the query's
current frame, and the two rows of a block differ only in the source of the
reference. Under the backbone the mass is spread over edges, ceiling lights
and background in both the reference and the query, and it changes little
between the two rows. Under ours it sits on the hand and the manipulated
object, in the reference and in the query alike, and it stays there when the
reference comes from a different source in a different place.

\subsection{Attention on Robot and Human Frames, Three Models}
\label{app:frames}

\begin{figure}[p]
    \centering
    \includegraphics[height=0.92\textheight]{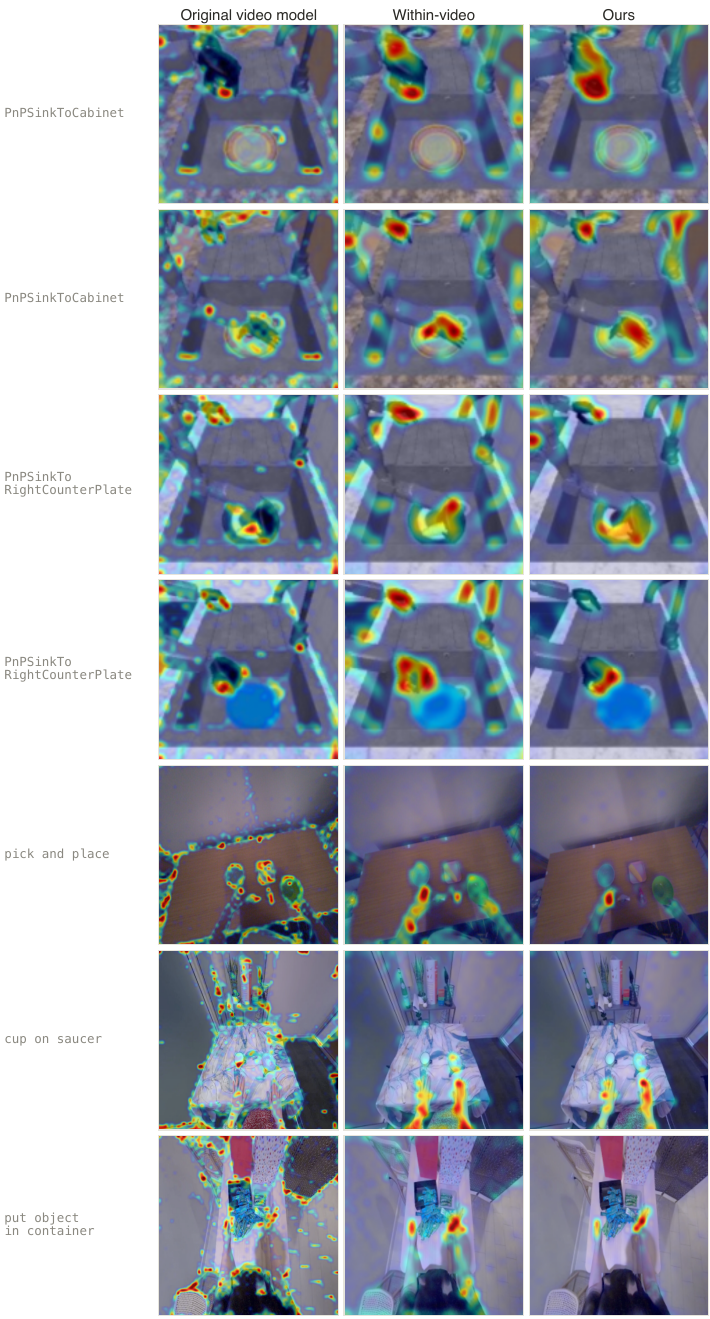}
    \caption{\textbf{Attention on robot and human frames, three models.}
    Attention to the current frame under the original video model
    (Cosmos-Predict2.5), the same model pre-trained within-video, and ours.
    Top four rows: side-right camera of GR1 MimicDroid rollouts of the task
    named on the left. Bottom three rows: three EgoVerse queries, on the
    query's current frame.}
    \label{fig:app-frames}
\end{figure}

Fig.~\ref{fig:app-frames} shows the current-frame attention of the original
video model, the within-video baseline and ours on four robot frames (top four
rows, side-right camera of GR1 MimicDroid rollouts) and three EgoVerse queries
(bottom three rows). The original video model spreads over edges and
background, within-video pre-training moves the mass onto the arm, and
cross-video pre-training puts it on the hand and the object.

\subsection{Analysis Protocol for the Representation Analyses}
\label{app:repr}

This subsection documents how the representation analyses of
\S\ref{sec:what-learns} (Fig.~\ref{fig:invariance}) and of
\S\ref{app:probe-more} were run: the models, the two window sets, the
features, and the retrieval and probing measures. It adds no new result.

\paragraph{Models.} The \emph{original video model} is the released
Cosmos-Predict2.5-2B image-to-video checkpoint, with no EgoVerse training. The
\emph{within-video} baseline and \emph{ours} are pre-trained on EgoVerse for
$8$K iterations with the same architecture and reference layout, within-video
with blank reference slots and ours with the retrieved reference.

\paragraph{Window sets.} The human set is the $359$ windows of the analysis set
(\S\ref{app:egoverse}): MECKA $169$, Scale $158$, Aria $32$. The robot
set is $396$ windows from RoboMIND~\citep{wu2025robomind} on three robots
(Franka, UR5e, TienKung) and six tasks (bread in basket, \rev{place object in basket, open lid},
open and close drawer, throw trash), every task on every robot, $22$ episodes
per cell and one window each (Fig.~\ref{fig:robomind-example}). The window
takes the midpoint of the motion path as the present, the halves before and
after as past state chunk and future, resampled to $92$ frames as for the human
windows; the camera is each robot's fixed top camera, and the wrist trajectory
is the recorded end-effector pose (Franka) or forward kinematics of the
recorded joints. No model has seen robot data.

\paragraph{Features.} Every window is read through each model in its training
layout: the current frame and the past state chunk as query, the reference
video slots blank for all three models, the text prompt naming the task and
the source or embodiment. \rev{For the retrieval and regression of
Fig.~\ref{fig:invariance}, all three models also receive the future actions of
the same matched reference from the same collection source, or the same robot,
in the retrieved-action slots, so that they differ only in their weights. The
classification probes of \S\ref{app:probe-more} give these actions to ours
alone, as in training, so the information in that reference can contribute to
its probing results alongside what pre-training changed in the features.} We capture the first
forward pass of generation (denoising step $0$, guidance $1.5$) and take the
spatial mean of the action-slot hidden state at each transformer layer, one
$2048$-dimensional vector per window and layer. Fig.~\ref{fig:invariance}
reads layers $10$--$27$, the later part of the $28$-layer network, where the
representation is closest to the action output, and averages over them
rather than picking one layer: retrieval ranks are averaged over these layers
per query, and probes are fit per layer and their scores averaged.

\paragraph{Retrieval.} A pair is a query and its matched partner from a
\emph{different} collection source for the human set ($120$ pairs, from the
training retrieval of \S\ref{app:egoverse}), or from a \emph{different}
robot for the robot set ($396$ pairs, by dynamic-time-warping
distance on size-normalized wrist trajectories within the same task). The
partner is ranked by cosine distance, on window-wise z-normalized features,
among all windows of the partner's source, or robot, and the rank is normalized to
$[0,1]$ with $0.5$ chance; the curve is the cumulative distribution of the
layer-averaged rank over the pairs. The area under the curve is \rev{$0.53$, $0.67$
and $0.79$ on the human set and $0.61$, $0.67$ and $0.73$ on the robot set} for
the original video model, within-video and ours.

\paragraph{Probes.} Labels come from the wrist displacement over the window, in
the waist frame for the human set and the robot base frame for the robot set.
Each of the three direction axes is a three-way label, positive beyond $+3$~cm,
still within $\pm3$~cm, negative beyond $-3$~cm (chance $1/3$); grasp is a
four-way label, open, grasp, release or hold, from the hand closure at the
first and last five frames of the window (pinch distance for the human set,
gripper opening for the robot set; chance $1/4$). The classifier is fit per
layer with standardization, PCA to $50$ components and logistic regression
under stratified five-fold cross-validation, scored by balanced accuracy; human
windows are probed per hand and averaged. Bars in Fig.~\ref{fig:probe-more}
are the mean over layers $10$--$27$ and error bars the standard error across
layers, not a sampling interval. With one window per episode, the robot set's
direction labels are tied to the task and robot, hence its high absolute
accuracies.

\begin{figure}[t]
    \centering
    \includegraphics[width=0.60\linewidth]{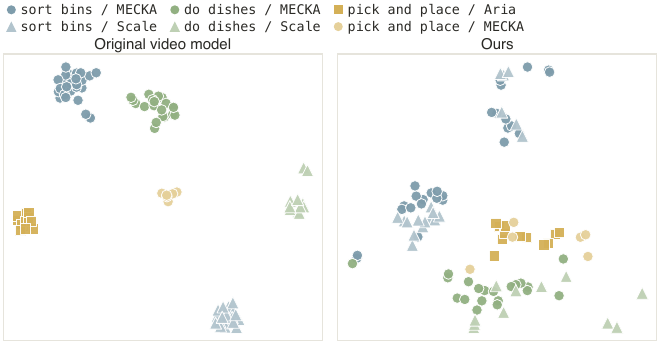}
    \caption{\textbf{Task layout across collection sources.} t-SNE of layer-$0$
    action-slot features for $95$ human windows of three tasks, each recorded
    by two of the three sources. Color is the task, in a full and a pale shade
    for its two sources, and shape is the source. Under the original video
    model each source forms its own island within a task; under ours the two
    sources of a task share an island.}
    \label{fig:tsne-large}
\end{figure}

\paragraph{t-SNE.} Fig.~\ref{fig:tsne-large} embeds layer-$0$ features into
two dimensions with t-SNE, fitted separately per model on the same windows, so
the panels share the points but not the axes. It draws the $95$ human windows
of three tasks each recorded by two sources, sort bins (MECKA $27$, Scale
$19$), do dishes (MECKA $20$, Scale $11$) and pick and place (Aria $12$, MECKA
$6$); single-source tasks are left out, since for them source and task cannot
be told apart, and the container task because its $62$ windows crowd the panel.
The embedding uses perplexity $25$ on $30$ principal components, is initialized
from the leading two components so it is deterministic, and runs $700$
iterations with early exaggeration over the first $100$.

\subsection{Additional Probing Results}
\label{app:probe-more}

\begin{figure}[t]
    \centering
    \includegraphics[width=\linewidth]{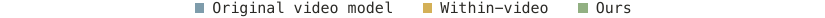}\\[0pt]
    \begin{subfigure}[t]{0.273\textwidth}
        \includegraphics[width=\linewidth]{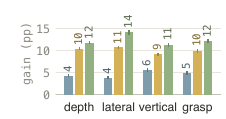}
        \caption{Probing, human}
    \end{subfigure}\hspace{0.03\textwidth}
    \begin{subfigure}[t]{0.244\textwidth}
        \includegraphics[width=\linewidth]{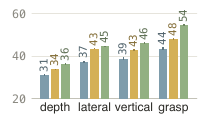}
        \caption{Probing, robot}
    \end{subfigure}\\[4pt]
    \begin{subfigure}[t]{0.347\textwidth}
        \includegraphics[width=\linewidth]{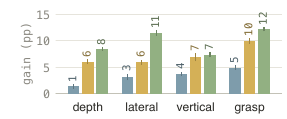}
        \caption{4-way probing, human}
    \end{subfigure}\hspace{0.03\textwidth}
    \begin{subfigure}[t]{0.318\textwidth}
        \includegraphics[width=\linewidth]{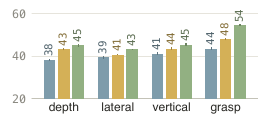}
        \caption{4-way probing, robot}
    \end{subfigure}
    \caption{\textbf{Linear probes of motion attributes.} Layers $10$--$27$,
    as in Fig.~\ref{fig:invariance}; bars are linear-probe balanced accuracy
    above each attribute's chance level, in percentage points, error bars the
    standard error across layers. (a, b) Three-way direction probes and the
    four-way grasp probe on human (a) and robot (b) windows. (c, d) Four-way
    probes: each direction axis is split by sign and, within each sign, at the
    median magnitude, so the label carries the direction and a coarse size;
    grasp is the four-state label as before (chance $1/4$ throughout).}
    \label{fig:probe-more}
\end{figure}

Fig.~\ref{fig:probe-more} probes the features of Fig.~\ref{fig:invariance}
for what they hold about the motion itself. The three-way probes (a, b)
improve under ours on all four attributes on both sets, with the largest gains
over the original video model in lateral direction for humans and grasp state
for robots. The four-way probes (c, d) ask for the size of the motion as well
as its direction; the ordering of the models is unchanged on every attribute,
and the gains of ours are again largest on lateral direction and grasp
(human) and on grasp (robot).

The held-out-embodiment regression of Fig.~\ref{fig:invariance}c asks
whether motion read from two embodiments transfers to a third. A kernel-ridge
regressor is trained on the features of two embodiments and tested on the
held-out one for nine quantities of the future motion (displacement, path
length, speeds, timing and contact terms), each z-scored within its
embodiment; the score is the Spearman correlation averaged over quantities
and the three folds. Ours transfers best on both sets, at $0.18$ on the human
set and \rev{$0.28$} on the robot set against \rev{$0.10$ and $0.03$ for the original
video model and $0.15$ and $0.25$ for within-video, with all three models
given the same reference actions (\S\ref{app:repr})}, and the robot-set gain over the original model is the larger one even though no model
has seen robot data. \rev{One caveat:} the Aria source has only $32$ windows, so one of the
three human folds is noisy.

\section{Human Data Against Robot Data as Pre-Training Data}
\label{app:robomind-mid}

\subsection{Setup}
\label{app:robomind-setup}

This section asks how close pre-training on human video comes to
pre-training on real robot data. We compare two arms with the same recipe and
a comparable amount of paired video: \method on a $10\%$ subset of
the human pairs, about $1$K hours, and the same objective on RoboMIND pairs
alone, also about $1$K hours, built with the same window and pairing definitions
and counted as in \S\ref{app:stats}. Only the robot arm sees RoboMIND during
pre-training, and both are compared on RoboMIND windows, so agreement between
them is the human arm matching a model trained on the test robots. The
comparison set is a RoboMIND subset of $300$ windows, $100$ Franka, $71$ UR5e
and $129$ TienKung, drawn so that the three robots share the same tasks.

\subsection{Analysis}
\label{app:robomind-analysis}

\begin{figure}[t]
    \centering
    \includegraphics[width=\linewidth]{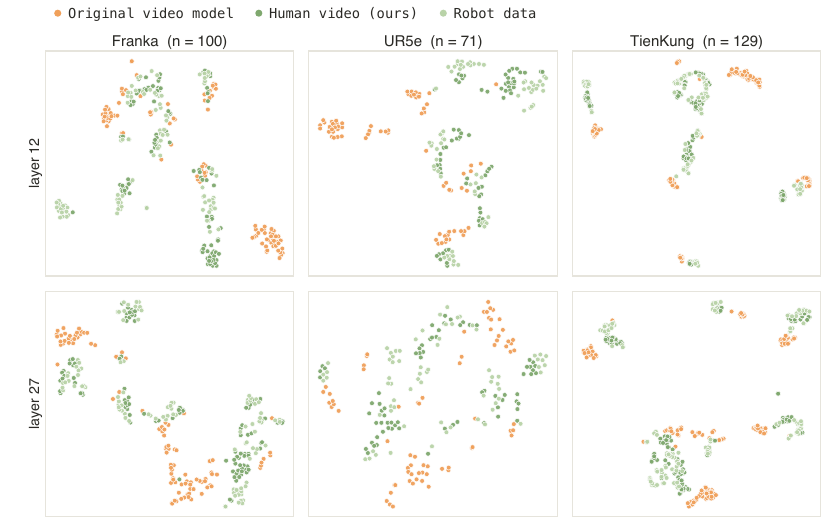}
    \caption{\textbf{Feature space of the human-trained and the robot-trained
    model on robot windows.} Action-slot features of $300$ RoboMIND windows,
    per robot, at layers $12$ and $27$: each model is reduced to $10$
    principal components, the other two are rotated onto the human-video
    arm's components (two-fold cross-fit), and the three are embedded jointly
    with t-SNE, so every window appears three times.}
    \label{fig:tsne-robomind}
\end{figure}

\paragraph{Feature space.} Fig.~\ref{fig:tsne-robomind} embeds the three
models' features of the same robot windows jointly, after each model's
feature space has been rotated onto the human arm's, so that a window's three
copies would coincide if the models agreed. The human-video and the
robot-data arm do agree: every cluster holds both greens, and a window's two
copies sit at a mean distance of $0.03$ to $0.06$ after alignment, against
$0.12$ to $0.31$ for the original video model, two to seven times farther
on every robot and at both layers. The original model's copies form
clusters of their own, so its disagreement is not noise around the same
layout but a different organization of the same frames. The agreement holds
across the three robots, on the three-camera Franka rig as on the single-camera
UR5e and TienKung, and at both depths of the network, at layer $12$ and at
layer $27$ next to the action head. A single objective applied to human pairs
thus moves the representation of robot frames to nearly where the same
objective applied to the robot data itself moves it, although the human arm
has never seen a robot.

\begin{figure}[t]
    \centering
    \includegraphics[width=\linewidth]{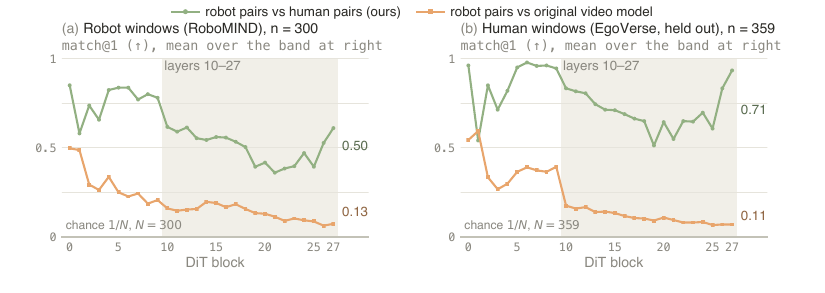}
    \caption{\textbf{Agreement with the robot-trained model, layer by
    layer.} Procrustes match@$1$ between the action-slot features of the
    robot-data arm and those of the human-video arm (green) or of the
    original video model (orange) at every DiT block, on (a) $300$ RoboMIND
    windows and (b) $359$ held-out EgoVerse windows. Each model's features
    are reduced to $50$ principal components and scale-normalized; one model
    is rotated onto the other by an orthogonal Procrustes map fitted on half
    the windows and applied to the other half, the halves swapping roles so
    that every window is placed by a map it did not contribute to. The score
    is the fraction of windows whose nearest neighbor among all $N$ reference
    points is the window itself, so chance is $1/N$: $0.0033$ in (a) and
    $0.0028$ in (b). Shaded: layers $10$--$27$, the band of the summaries,
    with its mean at the right.}
    \label{fig:match-layers}
\end{figure}

\paragraph{Agreement by layer.} Fig.~\ref{fig:match-layers} applies a
stricter test at every block: after a rotation fitted on other windows, a
window's nearest neighbor under the other model should still be the window
itself. The human-video arm keeps the robot-data arm's neighbors at every depth: on
robot windows match@$1$ stays between $0.36$ and $0.85$ across the $28$
blocks, with a mean of $0.50$ over layers $10$--$27$, and on human windows
between $0.51$ and $0.98$, with a mean of $0.71$, against a chance level of
$1/N$, at most $0.0033$ here. The original video model does not: against the robot arm it drops
to $0.20$ or less from layer $10$ on, with band means of $0.13$ and $0.11$;
the first ten blocks, where it still agrees at $0.2$ to $0.6$, carry the
input-side features all three models share. From the tenth block on, then, a
window keeps its neighbors across the human and the robot arm and loses them
against the model both started from.

\begin{figure}[t]
    \centering
    \begin{subfigure}{\linewidth}
        \centering
        \includegraphics[width=\linewidth]{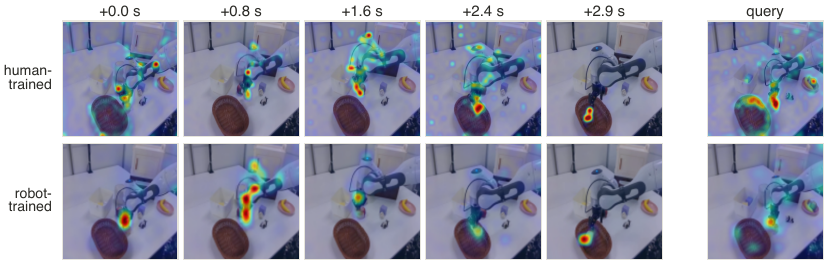}
        \caption{Bread in basket, Franka query}
    \end{subfigure}\\[3pt]
    \begin{subfigure}{\linewidth}
        \centering
        \includegraphics[width=\linewidth]{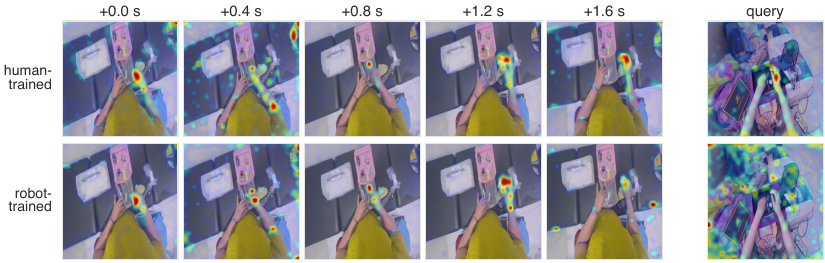}
        \caption{Sort bins, MECKA query}
    \end{subfigure}
    \caption{\textbf{Attention of the human-trained and the robot-trained
    model.} Action-token attention under a same-embodiment reference, five
    points of its span then the query frame. (a) A Franka query, bread in
    basket, from the left camera for both arms. (b) A MECKA query, sort bins,
    over the first half of the reference, where the sorting happens.}
    \label{fig:attention-robomind}
\end{figure}

\paragraph{Attention.} Fig.~\ref{fig:attention-robomind} asks the same
question of where the two arms look, on one query from each domain. On the
Franka query both arms follow the gripper and the bread through the
reference, shifting their mass from the bread on the table to the gripper as
it closes, and in the current frame both concentrate on the gripper and the
basket, the human arm on a robot it has never seen. On
the MECKA query both arms track the hand and the object it is sorting
through the reference and land on the hands in the current frame, the robot
arm now on a human scene it has never seen. Neither arm spreads its mass
over the background the way the original video model does
(Fig.~\ref{fig:motivation}b). In both directions, then, the arm trained
on one embodiment attends to the other's frames the way the arm trained on
that embodiment does.

\paragraph{Reading.} Together, these analyses show a correspondence between
the features learned from human pairs and from robot pairs, alongside similar
attention patterns around hands or grippers and the manipulated object. These
observations suggest that human-pair pre-training captures aspects of
interaction that robot demonstrations also teach.

\subsection{Downstream Transfer from Human and Robot Pre-Training}
\label{app:robomind-post}

\rev{Figure~\ref{fig:scale} places human-pair and robot-pair pre-training on one collection-budget axis, with the MimicDroid post-training protocol and target-robot data held fixed.
An arm's budget is its amount of paired video times the cost of collecting a unit of that video, normalized by the estimated cost of the full human corpus, following the relative collection-cost analysis of RoboVQA~\citep{sermanet2024robovqa}.
We do not measure the robot-to-human cost ratio, so we use $8\times$ as the central setting and show $4\times$ to $16\times$.
This range follows reported collection throughput: human demonstrations are collected at $1{,}400$ against $135$ robot-teleoperation demonstrations per hour in EgoMimic~\citep{kareer2025egomimic}, about $10\times$, and at $4.6\times$ the rate of a Gello teleoperation setup in DexWild~\citep{tao2025dexwild}, while even a handheld gripper interface is more than $3\times$ faster than teleoperation~\citep{chi2024universal}. These rates count operator time only and leave out the robot hardware itself.
Each robot line ends at the largest robot setting we ran, and both arms start at $0$ from the same model without this pre-training stage.}


\section{Additional Experimental Results \& Details}
\label{app:results}

\subsection{Baselines}
\label{app:baselines}

All baselines we train are post-trained on the same robot demonstrations as
\method (\S\ref{sec:setup}). MimicDroid is evaluated over $30$ episodes per task ($40$ for the L2 plate
task, \S\ref{app:mimicdroid}), TableTop-Sim and RoboCasa over $50$
episodes per task. The vision-language-action baselines
$\pi_{0.5}$~\citep{pi05} and MolmoAct2~\citep{fang2026molmoact2actionreasoningmodels}
cap the action dimension their heads accept, so on MimicDroid they use a
$20$-dimensional action vector rather than the $57$-d representation of
\S\ref{app:representation}, which every other model uses there; on
TableTop-Sim every model uses ALOHA's $20$-d action target. GR00T-N1.7~\citep{nvidia2025gr00t} and
Vid2Robot~\citep{jain2024vid2robot} follow their released recipes on the same
data; HAND~\citep{hong2025hand} and MimicDroid~\citep{shah2025mimicdroid} are
reported by prior work under its own evaluation data.

\newcommand{\taskhead}[2]{\iflatexml #2\else\rotatebox{90}{\sffamily\scriptsize #1}\fi}
\newcommand{\taskheadlegend}{\iflatexml\ Task abbreviations: CLD/CRD = Close Left/Right CabinetDoor, Faucet = TurnOnFaucet, S2Cab = PnPSinkToCabinet, S2Plate = PnPSinkToRightCounterPlate, S2MW = PnPSinkToMicrowaveTop.\fi}

\subsection{Per-Task Results}
\label{app:per-task}

\begin{table}[t]
\centering
\caption{\textbf{In-context MimicDroid, per task.} Success rate (\%) of every
model of Table~\ref{tab:icl-main} and Table~\ref{tab:ablation} on the four
tasks of each level; \textit{Avg.} is the mean over the tasks of a level, and
the last column the mean over L1--L3. Best in bold, second best underlined.\taskheadlegend}
\label{tab:per-task-icl}
\newcommand{\grouprowpt}[1]{\multicolumn{17}{l}{\rule{0pt}{6.5pt}\textcolor{gray}{\scriptsize\textit{#1}}} \\[-1.5pt]}
\scriptsize
\setlength{\tabcolsep}{2.2pt}
\renewcommand{\arraystretch}{1.0}
\begin{tabular}{l|ccccc|ccccc|ccccc|c}
    \toprule
    \multirow{2}{*}{Method} & \multicolumn{5}{c|}{L1} & \multicolumn{5}{c|}{L2} & \multicolumn{5}{c|}{L3} & \\
    & \taskhead{CloseLeftCabinetDoor}{CLD} & \taskhead{TurnOnFaucet}{Faucet} & \taskhead{PnPSinkToCabinet}{S2Cab} & \taskhead{PnPSinkToRightCounterPlate}{S2Plate} & \taskhead{Avg.}{Avg.} & \taskhead{CloseLeftCabinetDoor}{CLD} & \taskhead{CloseRightCabinetDoor}{CRD} & \taskhead{PnPSinkToCabinet}{S2Cab} & \taskhead{PnPSinkToRightCounterPlate}{S2Plate} & \taskhead{Avg.}{Avg.} & \taskhead{CloseLeftCabinetDoor}{CLD} & \taskhead{TurnOnFaucet}{Faucet} & \taskhead{PnPSinkToMicrowaveTop}{S2MW} & \taskhead{PnPSinkToRightCounterPlate}{S2Plate} & \taskhead{Avg.}{Avg.} & \taskhead{Avg.}{Avg.} \\
    \midrule
    \grouprowpt{ICL baselines}
    Vid2Robot~\citep{jain2024vid2robot} & \textbf{100.0} & \textbf{100.0} & 63.3 & \textbf{100.0} & \underline{90.8} & 53.3 & 0.0 & 63.3 & \textbf{87.5} & 51.0 & \underline{0.0} & 0.0 & 0.0 & 3.3 & 0.8 & 47.6 \\
    \grouprowpt{VLA baselines}
    $\pi_{0.5}$~\citep{pi05} & \textbf{100.0} & \textbf{100.0} & 80.0 & 36.7 & 79.2 & 43.3 & 0.0 & 70.0 & 40.0 & 38.3 & \textbf{33.3} & 0.0 & 0.0 & 0.0 & 8.3 & 41.9 \\
    GR00T-N1.7~\citep{nvidia2025gr00t} & \textbf{100.0} & \textbf{100.0} & 33.3 & 80.0 & 78.3 & \textbf{100.0} & 0.0 & 66.7 & 70.0 & 59.2 & \textbf{33.3} & 6.7 & 0.0 & 0.0 & 10.0 & 49.2 \\
    \grouprowpt{Cosmos-Predict2.5-2B}
    Cosmos Policy~\citep{kim2026cosmospolicy} & \textbf{100.0} & \textbf{100.0} & 40.0 & 60.0 & 75.0 & 0.0 & 33.3 & 6.7 & 77.5 & 29.4 & \underline{0.0} & \underline{70.0} & 0.0 & 3.3 & 18.3 & 40.9 \\
    \quad + Within-video & \textbf{100.0} & \textbf{100.0} & 63.3 & 80.0 & 85.8 & \textbf{100.0} & 0.0 & 40.0 & 37.5 & 44.4 & \textbf{33.3} & 56.7 & 0.0 & 3.3 & 23.3 & 51.2 \\
    \quad + Co-training & \textbf{100.0} & \textbf{100.0} & 53.3 & \underline{96.7} & 87.5 & \textbf{100.0} & 33.3 & 70.0 & 30.0 & 58.3 & \textbf{33.3} & 66.7 & 0.0 & 36.7 & 34.2 & 60.0 \\
    \quad + \method, same-source top-1 & \textbf{100.0} & \textbf{100.0} & 70.0 & 66.7 & 84.2 & \underline{86.7} & 0.0 & 66.7 & \underline{80.0} & 58.3 & \textbf{33.3} & 66.7 & 0.0 & 33.3 & 33.3 & 58.6 \\
    \rowcolor{ngreenlight}
    \quad + \method (Ours) & \textbf{100.0} & \underline{86.7} & \textbf{100.0} & 80.0 & \textbf{91.7} & 66.7 & \underline{66.7} & \textbf{86.7} & 70.0 & 72.5 & \textbf{33.3} & 66.7 & \underline{30.0} & 33.3 & \underline{40.8} & \underline{68.3} \\
    \grouprowpt{Cosmos3 Edge-2B}
    Cosmos3~\citep{nvidia2026cosmos3} & \textbf{100.0} & \textbf{100.0} & 46.7 & 30.0 & 69.2 & \textbf{100.0} & \textbf{100.0} & 60.0 & 57.5 & \underline{79.4} & \textbf{33.3} & 0.0 & 6.7 & \textbf{83.3} & 30.8 & 59.8 \\
    \rowcolor{ngreenlight}
    \quad + \method (Ours) & \underline{86.7} & \textbf{100.0} & \underline{90.0} & 90.0 & \textbf{91.7} & \textbf{100.0} & \textbf{100.0} & \underline{73.3} & 72.5 & \textbf{86.5} & \textbf{33.3} & \textbf{80.0} & \textbf{40.0} & \underline{66.7} & \textbf{55.0} & \textbf{77.7} \\
    \bottomrule
\end{tabular}
\end{table}

\paragraph{In-context MimicDroid.} Table~\ref{tab:per-task-icl} breaks the in-context MimicDroid numbers of
Tables~\ref{tab:icl-main} and~\ref{tab:ablation} down by task. L1 is saturated
on the door and faucet tasks for nearly every model, so the level is decided
by the two pick-and-place tasks; on L2 and L3 cross-video
pre-training improves multiple tasks at each level, and only the cross-video models
reach double digits on L3 {\small\texttt{PnPSinkToMicrowaveTop}}.

\begin{table}[t]
\centering
\caption{\textbf{Context-free MimicDroid, per task.} Success rate (\%) of every
model of Table~\ref{tab:bc-main} and Table~\ref{tab:ablation} on the four L1
tasks and the four L2 tasks; \textit{Avg.} is the mean over the tasks of a
level, and the last column the mean over the two levels. Best in bold, second
best underlined.\taskheadlegend}
\label{tab:per-task}
\newcommand{\grouprowcf}[1]{\multicolumn{12}{l}{\rule{0pt}{6.5pt}\textcolor{gray}{\scriptsize\textit{#1}}} \\[-1.5pt]}
\footnotesize
\setlength{\tabcolsep}{2.4pt}
\renewcommand{\arraystretch}{1.0}
\begin{tabular}{l|ccccc|ccccc|c}
    \toprule
    \multirow{2}{*}{Method} & \multicolumn{5}{c|}{L1} & \multicolumn{5}{c|}{L2} & \\
    & \taskhead{CloseLeftCabinetDoor}{CLD} & \taskhead{TurnOnFaucet}{Faucet} & \taskhead{PnPSinkToCabinet}{S2Cab} & \taskhead{PnPSinkToRightCounterPlate}{S2Plate} & \taskhead{Avg.}{Avg.} & \taskhead{CloseLeftCabinetDoor}{CLD} & \taskhead{CloseRightCabinetDoor}{CRD} & \taskhead{PnPSinkToCabinet}{S2Cab} & \taskhead{PnPSinkToRightCounterPlate}{S2Plate} & \taskhead{Avg.}{Avg.} & \taskhead{Avg.}{Avg.} \\
    \midrule
    \grouprowcf{VLA baselines}
    $\pi_{0.5}$~\citep{pi05} & \textbf{100.0} & \textbf{100.0} & 73.3 & 40.0 & 78.3 & 10.0 & \textbf{0.0} & 56.7 & 52.5 & 29.8 & 54.1 \\
    GR00T-N1.7~\citep{nvidia2025gr00t} & \textbf{100.0} & \textbf{100.0} & 43.3 & 73.3 & 79.2 & \underline{76.7} & \textbf{0.0} & 56.7 & 50.0 & \underline{45.8} & 62.5 \\
    MolmoAct2~\citep{fang2026molmoact2actionreasoningmodels} & \textbf{100.0} & \textbf{100.0} & 26.7 & 80.0 & 76.7 & 70.0 & \textbf{0.0} & 3.3 & 40.0 & 28.3 & 52.5 \\
    \grouprowcf{Cosmos-Predict2.5-2B}
    Cosmos Policy~\citep{kim2026cosmospolicy} & \textbf{100.0} & 43.3 & 53.3 & \underline{90.0} & 71.7 & 0.0 & \textbf{0.0} & 30.0 & 77.5 & 26.9 & 49.3 \\
    \quad + Within-video pre-training & \textbf{100.0} & \textbf{100.0} & 63.3 & \underline{90.0} & 88.3 & 0.0 & \textbf{0.0} & 83.3 & \underline{80.0} & 40.8 & 64.6 \\
    \quad + Co-training & \textbf{100.0} & \underline{93.3} & 73.3 & \textbf{93.3} & \underline{90.0} & 0.0 & \textbf{0.0} & \textbf{96.7} & 72.5 & 42.3 & \underline{66.1} \\
    \quad + \method, same-source top-1 & \textbf{100.0} & \textbf{100.0} & \textbf{100.0} & 60.0 & \underline{90.0} & 0.0 & \textbf{0.0} & 66.7 & \underline{80.0} & 36.7 & 63.3 \\
    \rowcolor{ngreenlight}
    \quad + \method (Ours) & \textbf{100.0} & \textbf{100.0} & 76.7 & 86.7 & \textbf{90.8} & 0.0 & \textbf{0.0} & \underline{93.3} & \textbf{95.0} & \textbf{47.1} & \textbf{69.0} \\
    \grouprowcf{Cosmos3 Edge-2B}
    Cosmos3~\citep{nvidia2026cosmos3} & \underline{80.0} & \textbf{100.0} & 80.0 & 53.3 & 78.3 & 0.0 & \textbf{0.0} & 26.7 & 57.5 & 21.0 & 49.7 \\
    \rowcolor{ngreenlight}
    \quad + \method (Ours) & \textbf{100.0} & \textbf{100.0} & \underline{93.3} & 70.0 & \textbf{90.8} & 0.0 & \textbf{0.0} & 66.7 & 77.5 & 36.0 & 63.4 \\
    \bottomrule
\end{tabular}
\end{table}

\paragraph{Context-free MimicDroid.} Table~\ref{tab:per-task} breaks the context-free MimicDroid numbers of
Tables~\ref{tab:bc-main} and~\ref{tab:ablation} down by task (the cabinet
tasks close the doors).
The two L2 cabinet doors score near zero for most models, so the L2 average
is driven by the pick-and-place tasks, where cross-video pre-training gains
most.

\begin{table}[t]
\centering
\caption{\textbf{TableTop-Sim, per task.} Success rate (\%) of every model of
Table~\ref{tab:bc-main} and Table~\ref{tab:ablation} on the five tasks of the
easy (seen) and the hard (unseen) split under context-free post-training;
\textit{Avg.} is the mean over the tasks of a split, and the last column the
mean over the two splits. Best in bold, second best underlined.}
\label{tab:per-task-tt}
\newcommand{\grouprowtt}[1]{\multicolumn{14}{l}{\rule{0pt}{6.5pt}\textcolor{gray}{\scriptsize\textit{#1}}} \\[-1.5pt]}
\footnotesize
\setlength{\tabcolsep}{2.2pt}
\renewcommand{\arraystretch}{1.0}
\begin{tabular}{l|cccccc|cccccc|c}
    \toprule
    \multirow{2}{*}{Method} & \multicolumn{6}{c|}{Easy (seen)} & \multicolumn{6}{c|}{Hard (unseen)} & \\
    & \taskhead{handover box}{handover box} & \taskhead{lift box}{lift box} & \taskhead{shoes table}{shoes table} & \taskhead{dish drainer}{dish drainer} & \taskhead{box into pot}{box into pot} & \taskhead{Avg.}{Avg.} & \taskhead{handover box}{handover box} & \taskhead{lift box}{lift box} & \taskhead{shoes table}{shoes table} & \taskhead{dish drainer}{dish drainer} & \taskhead{box into pot}{box into pot} & \taskhead{Avg.}{Avg.} & \taskhead{Avg.}{Avg.} \\
    \midrule
    \grouprowtt{VLA baselines}
    $\pi_{0.5}$~\citep{pi05} & 66.0 & 42.0 & 80.0 & 90.0 & 74.0 & 70.4 & 48.0 & 0.0 & 44.0 & 56.0 & 22.0 & 34.0 & 52.2 \\
    GR00T-N1.7~\citep{nvidia2025gr00t} & 62.0 & 30.0 & 84.0 & 84.0 & 74.0 & 66.8 & 56.0 & 14.0 & \textbf{82.0} & 76.0 & 26.0 & 50.8 & 58.8 \\
    MolmoAct2~\citep{fang2026molmoact2actionreasoningmodels} & 70.0 & 32.0 & 70.0 & 92.0 & 68.0 & 66.4 & 20.0 & \textbf{22.0} & 42.0 & 86.0 & 28.0 & 39.6 & 53.0 \\
    \grouprowtt{Cosmos-Predict2.5-2B}
    Cosmos Policy~\citep{kim2026cosmospolicy} & \underline{82.0} & 34.0 & 72.0 & 78.0 & \underline{82.0} & 69.6 & 76.0 & 14.0 & 60.0 & 74.0 & 24.0 & 49.6 & 59.6 \\
    \quad + Within-video & 68.0 & 34.0 & 74.0 & \textbf{100.0} & \underline{82.0} & 71.6 & 52.0 & \underline{20.0} & 62.0 & \textbf{100.0} & 24.0 & 51.6 & 61.6 \\
    \quad + Co-training & 74.0 & 34.0 & 76.0 & 92.0 & 70.0 & 69.2 & 76.0 & 14.0 & \underline{66.0} & 86.0 & 22.0 & 52.8 & 61.0 \\
    \quad + \method, same-source top-1 & 62.0 & 44.0 & 76.0 & 94.0 & 72.0 & 69.6 & 62.0 & 10.0 & 62.0 & \underline{92.0} & 30.0 & 51.2 & 60.4 \\
    \rowcolor{ngreenlight}
    \quad + \method (Ours) & \underline{82.0} & 40.0 & 76.0 & 96.0 & \textbf{90.0} & \underline{76.8} & 54.0 & \textbf{22.0} & \textbf{82.0} & 86.0 & 26.0 & 54.0 & 65.4 \\
    \grouprowtt{Cosmos3 Edge-2B}
    Cosmos3~\citep{nvidia2026cosmos3} & 78.0 & \underline{46.0} & \underline{88.0} & \underline{98.0} & 72.0 & 76.4 & \underline{78.0} & 0.0 & 54.0 & 72.0 & \underline{72.0} & \underline{55.2} & \underline{65.8} \\
    \rowcolor{ngreenlight}
    \quad + \method (Ours) & \textbf{96.0} & \textbf{58.0} & \textbf{92.0} & 92.0 & \textbf{90.0} & \textbf{85.6} & \textbf{92.0} & 0.0 & 44.0 & 70.0 & \textbf{86.0} & \textbf{58.4} & \textbf{72.0} \\
    \bottomrule
\end{tabular}
\end{table}

\paragraph{TableTop-Sim.} Table~\ref{tab:per-task-tt} breaks the TableTop-Sim
numbers of Tables~\ref{tab:bc-main} and~\ref{tab:ablation} down by task.
Lift box is the hard case on both splits; box into pot on the unseen split is
where the backbones part, Cosmos3 keeping its seen performance while the
Cosmos Policy variants fall sharply.

\subsection{Data Scaling Setup}
\label{app:ladder}

The sweep of \S\ref{sec:scaling} draws $25$, $50$ and $75\%$ of the query
episodes. The split is made within each collection source and rounded up, so
the arms hold $3{,}330$, $6{,}659$ and $9{,}988$ of the $13{,}315$ episodes
that carry query windows (\S\ref{app:stats}), a few episodes above the exact
quarters, and $25.6$, $49.7$ and $74.4\%$ of the query windows. Training length scales with the data, $2$K,
$4$K, $6$K and $8$K iterations at a global batch of $64$, so every arm sees the
same $0.49$ epochs; data and compute therefore grow together, and the sweep
scales the corpus at a fixed number of epochs.

\subsection{Supervision Ablation: Post-Training Target as Well}
\label{app:ablation-full}

\begin{figure}[t]
    \centering
    \includegraphics[width=\linewidth]{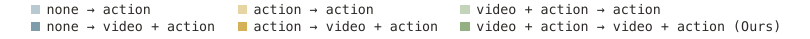}\\[0pt]
    \begin{subfigure}[t]{0.645\textwidth}
        \includegraphics[width=\linewidth]{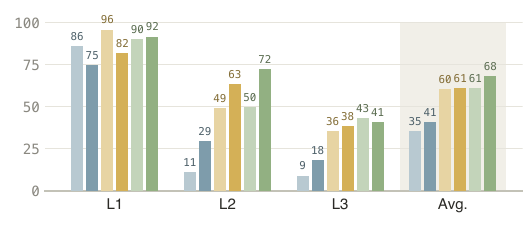}
        \caption{MimicDroid, in-context}
    \end{subfigure}\hfill
    \begin{subfigure}[t]{0.345\textwidth}
        \includegraphics[width=\linewidth]{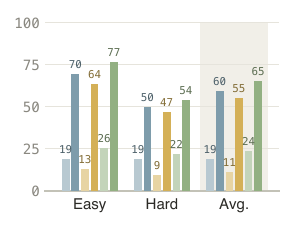}
        \caption{TableTop-Sim, context-free}
    \end{subfigure}
    \caption{\textbf{Supervision ablation, all configurations.} Success rate
    (\%). Hue: pre-training target (none, action, video\,+\,action); pale:
    action-only post-training, full: video\,+\,action. (a) MimicDroid,
    in-context, the reference given as actions only under action-only
    post-training; (b) TableTop-Sim, context-free.}
    \label{fig:ablation-full}
\end{figure}

Fig.~\ref{fig:ablation} fixes post-training at video\,+\,action and varies
what pre-training supervises; Fig.~\ref{fig:ablation-full} varies the
post-training target as well, so every bar reads \emph{pre-training target
$\rightarrow$ post-training target}, and under action-only post-training the
in-context reference is given as actions only. Video\,+\,action pre-training
is the best pre-training target under either post-training target on both
benchmarks. Action-only post-training is competitive on L1 and close on L3
but trails on L2 under every pre-training target, and collapses on
TableTop-Sim, which has no reference. Action-only pre-training helps on
MimicDroid, where the reference actions carry the task, but falls below no
pre-training on TableTop-Sim under both targets ($11$ and $55$ against $19$
and $60$).

\subsection{Robustness to Tracking Noise and Task Labels}
\label{app:label-noise}

\begin{table}[t]
\centering
\caption{\rev{\textbf{Raw hand tracking and RGB-only estimates.} Top: raw
tracking before preprocessing, median\,/\,90th percentile over the $13{,}449$
human training episodes; jitter is the RMS difference between the raw track and
its preprocessed version (\S\ref{app:representation}). Bottom: error of smoothed
HaWoR~\citep{zhang2025hawor} estimates against the tracked trajectories after
offset removal, median over $1{,}076$ three-second windows of the per-window
median error.}}
\label{tab:tracking-noise}
\small
\setlength{\tabcolsep}{7pt}
\renewcommand{\arraystretch}{1.1}
\begin{tabular}{l|ccc}
    \toprule
    & MECKA ($6{,}718$) & Scale ($6{,}074$) & Aria ($657$) \\
    \midrule
    wrist position jitter (mm)   & 3.1\,/\,7.9 & 1.6\,/\,4.5 & 14.4\,/\,27.4 \\
    fingertip jitter (mm)        & 3.6\,/\,9.0 & 2.0\,/\,4.9 & 15.7\,/\,30.1 \\
    dropped frames               & 0\%\,/\,0\% & 2.7\%\,/\,20.3\% & 0.9\%\,/\,2.5\% \\
    \midrule
    HaWoR wrist position (mm)    & 21 & 24 & 45 \\
    HaWoR wrist rotation (\textdegree) & 7 & 6 & 11 \\
    HaWoR fingertip position (mm) & 16 & 16 & 18 \\
    \bottomrule
\end{tabular}
\end{table}

\rev{\paragraph{Tracking noise.} The corpus spans capture pipelines with
different noise (Table~\ref{tab:tracking-noise}): Aria tracks show the largest
jitter and Scale tracks the most dropped frames, and together they make up
half of the episodes. In every source, the pairing does not degrade as the
noise grows: across noise quintiles, cross-source coverage does not drop and
the best DTW distance shows no trend with the noise (Spearman $|\rho| \le 0.03$
for Aria and Scale). This follows from the descriptor of
\S\ref{app:pairs}, which is standardized per window and compared by the shape
of the trajectory.}

\rev{\paragraph{RGB-based motion estimates.} To gauge whether hands estimated
from RGB alone could replace recorded tracking, we compare smoothed
HaWoR~\citep{zhang2025hawor} estimates with the tracked trajectories over
$1{,}076$ three-second windows ($2{,}072$ hand-windows). After offset removal,
the median wrist error is $21$--$45$\,mm and $6$--$11^\circ$, and the fingertip
error $16$--$18$\,mm, across the three sources (Table~\ref{tab:tracking-noise}).
We then select references with each motion source on $50$ queries, with $21$
candidate references per query on average, and score every selected pair on
the tracked trajectories, so the scores do not depend on the estimates used
for selection. Selection with HaWoR motion gives a speed-profile correlation
of $0.125$ [$0.064$, $0.191$] and a trajectory-shape cosine of $0.206$
[$0.118$, $0.293$], against $0.067$ [$0.006$, $0.125$] and $0.202$ [$0.122$,
$0.283$] with tracked motion and $0.015$ and $0.001$ for random selection
($95\%$ confidence intervals). Within this small evaluation, RGB-based selection
shows no loss of pair quality. The $50$-query set is separate from the
$3{,}000$-query set of Table~\ref{tab:label-noise-pairing}, so absolute values
are not comparable across the two.}

\rev{\paragraph{Pre-training on RGB-estimated hands.} We then replace the
tracked hands of every training episode with smoothed HaWoR estimates and use
them for everything that relied on tracking: pairing, the action targets, and
the reference actions. Task labels, backbone, and recipe are unchanged (last
row of Table~\ref{tab:label-noise-downstream}). Without tracking hardware, \method
still exceeds within-video pre-training on recorded tracking ($59.4$ against
$51.2\%$ in context, $62.8$ against $61.6\%$ on TableTop-Sim). Since the
estimates enter pairing, action targets, and reference actions together, the
experiment does not isolate their individual effects. The estimates deviate from the tracking by far more than the
tracking jitter (Table~\ref{tab:tracking-noise}), so this also measures the
effect of noisy action targets and reference actions; the drop concentrates
on in-context L2 and L3, where the policy must follow the reference's
motion.}

\paragraph{\rev{Task labels.}}

\begin{table}[t]
\centering
\caption{\rev{\textbf{Pairing under confused task labels.} Agreement between a
query and its top-ranked cross-source reference, scored against the true
labels with the metrics of Table~\ref{tab:pairing}, after re-running the
matcher on $3{,}000$ training queries with corrupted labels.}}
\label{tab:label-noise-pairing}
\small
\setlength{\tabcolsep}{7pt}
\renewcommand{\arraystretch}{1.1}
\begin{tabular}{l|cc}
    \toprule
    Labels used for pairing & speed-profile corr.\ $\uparrow$ & trajectory-shape cos.\ $\uparrow$ \\
    \midrule
    original                              & 0.20 & 0.33 \\
    $25\%$ confused with a related class  & 0.20 & 0.32 \\
    $50\%$ confused with a related class  & 0.20 & 0.33 \\
    super-group only ($10$ classes)       & 0.20 & 0.34 \\
    random pairs (null)                   & 0.00 & 0.01 \\
    \bottomrule
\end{tabular}
\end{table}

\begin{table}[t]
\centering
\caption{\rev{\textbf{Downstream success under confused task labels and
RGB-estimated hands.} Success rate (\%) on the Cosmos Policy backbone. The
fourth row rebuilds every pre-training pair with $50\%$ of the fine-grained
labels confused; the last row replaces the tracked hands of every training
episode with HaWoR estimates from RGB, for pairing, action targets, and
reference actions. Both keep the pre-training and post-training recipe of
Table~\ref{tab:ablation}.}}
\label{tab:label-noise-downstream}
\small
\setlength{\tabcolsep}{5pt}
\renewcommand{\arraystretch}{1.1}
\begin{tabular}{l|cccc|ccc}
    \toprule
    & \multicolumn{4}{c|}{MimicDroid, in-context} & \multicolumn{3}{c}{TableTop-Sim, context-free} \\
    Pre-training & L1 & L2 & L3 & Avg. & Easy & Hard & Avg. \\
    \midrule
    None                              & 75.0 & 29.4 & 18.3 & 40.9 & 69.6 & 49.6 & 59.6 \\
    Within-video                      & 85.8 & 44.4 & 23.3 & 51.2 & 71.6 & 51.6 & 61.6 \\
    \method, original labels          & 91.7 & \textbf{72.5} & \textbf{40.8} & \textbf{68.3} & \textbf{76.8} & 54.0 & \textbf{65.4} \\
    \method, $50\%$ confused labels   & 93.3 & 61.9 & 33.3 & 62.8 & 73.2 & \textbf{56.0} & 64.6 \\
    \method, RGB-estimated hands      & \textbf{94.2} & 50.6 & 33.3 & 59.4 & 73.6 & 52.0 & 62.8 \\
    \bottomrule
\end{tabular}
\end{table}

\rev{Task labels only gate the candidates of the pairing (\S\ref{app:pairs});
the motion matching ranks references within them. To test how much the
pairing depends on accurate labels, we replace a fraction of the
fine-grained labels with another class of the same super-group, drawn in
proportion to class frequency, and re-run the cross-source matcher.
Table~\ref{tab:label-noise-pairing} shows that motion agreement is unchanged
even with half of the labels confused, or with only the ten super-group
labels, and stays far above random pairs: it comes from the motion matching
rather than from the label. The values differ slightly from
Table~\ref{tab:pairing}, which uses $12{,}000$ queries.}

\rev{We then rebuild the full set of pre-training pairs with $50\%$ of the
labels confused and repeat pre-training and post-training
(Table~\ref{tab:label-noise-downstream}). \method keeps most of its gain and
remains above within-video pre-training on both benchmarks ($62.8$ against
$51.2\%$ in context, $64.6$ against $61.6\%$ on TableTop-Sim). TableTop-Sim,
which uses no reference at test time, is nearly unchanged; the drop
concentrates on in-context L2 and L3, where the policy must follow a reference
of the same task. Coarse labels therefore suffice, and for about half of the
corpus (MECKA) they are already derived automatically from free-form task
descriptions.}

\subsection{Real-Robot Evaluation}
\label{app:real}

\begin{figure}[t]
\centering
\includegraphics[width=\linewidth]{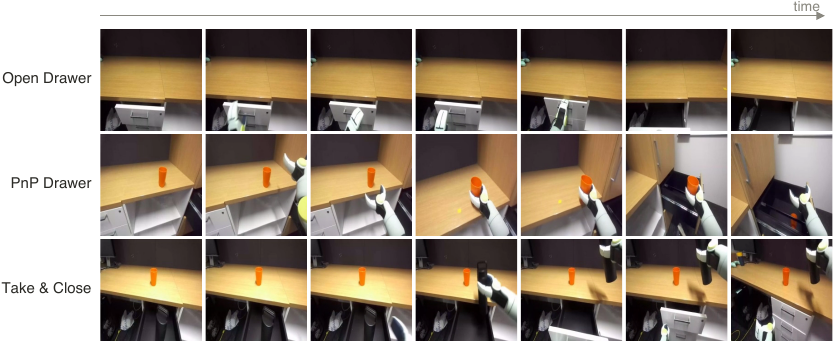}
\caption{\rev{\textbf{Real-robot post-training data.} One teleoperated
demonstration per task from the egocentric camera the policy uses
($224\times224$, $15$\,Hz), with time running left to right.}}
\label{fig:real-ego}
\end{figure}

\rev{\paragraph{Robot and data.} The robot is a humanoid~\citep{robotics2026fauna} with two arms,
parallel grippers, and an egocentric stereo camera, of which the policy uses
the left view at $224\times224$. Demonstrations are collected by
teleoperation at $15$\,Hz: $20$ per task, $60$ episodes and $11.5$ minutes in
total. The $29$-d action, expressed in the torso frame, holds a base velocity
command ($9$-d) and, for each arm, the end-effector position and $6$-d
rotation with a gripper command ($2\times10$-d); an arm a task does not use is
held at its reset pose. The robot moves its body during the tasks, so the
policy controls the whole body. Fig.~\ref{fig:real-ego} shows post-training demonstrations from the
egocentric camera the policy uses.}

\begin{figure}[t]
\centering
\includegraphics[width=\linewidth]{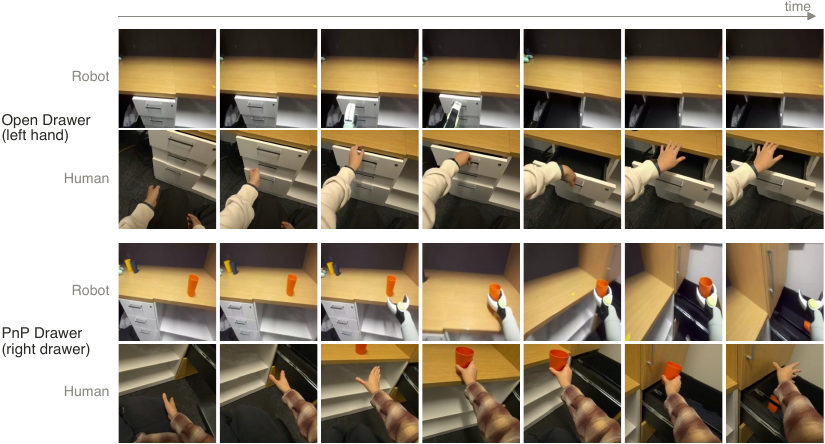}
\caption{\rev{\textbf{Real-robot in-context post-training data.} For each of
the two post-training tasks, a robot demonstration from the egocentric camera the
policy uses (top) and a human video of the same task paired with it as the
reference (bottom), with time running left to right. The two are recorded
separately and not time-aligned.}}
\label{fig:real-icl-data}
\end{figure}

\begin{table}[t]
\centering
\caption{\rev{\textbf{Real humanoid, per setting.} Success rate (\%) over
$10$ trials per condition, context-free (Fig.~\ref{fig:real}).}}
\label{tab:real}
\small
\setlength{\tabcolsep}{5pt}
\renewcommand{\arraystretch}{1.1}
\begin{tabular}{l|cc|cc|cc|c}
    \toprule
    & \multicolumn{2}{c|}{Open Drawer} & \multicolumn{2}{c|}{PnP Drawer} & \multicolumn{2}{c|}{Take \& Close} & \\
    Pre-training & Easy & Hard & Easy & Hard & Easy & Hard & Avg. \\
    \midrule
    None & 90 & 80 & 40 & 40 & 60 & 10 & 53.3 \\
    Within-video & \textbf{100} & \textbf{100} & 60 & \textbf{60} & 70 & 10 & 66.7 \\
    \method (Ours) & \textbf{100} & \textbf{100} & \textbf{80} & \textbf{60} & \textbf{90} & \textbf{50} & \textbf{80.0} \\
    \bottomrule
\end{tabular}
\end{table}

\begin{figure}[t]
\centering
\includegraphics[width=\linewidth]{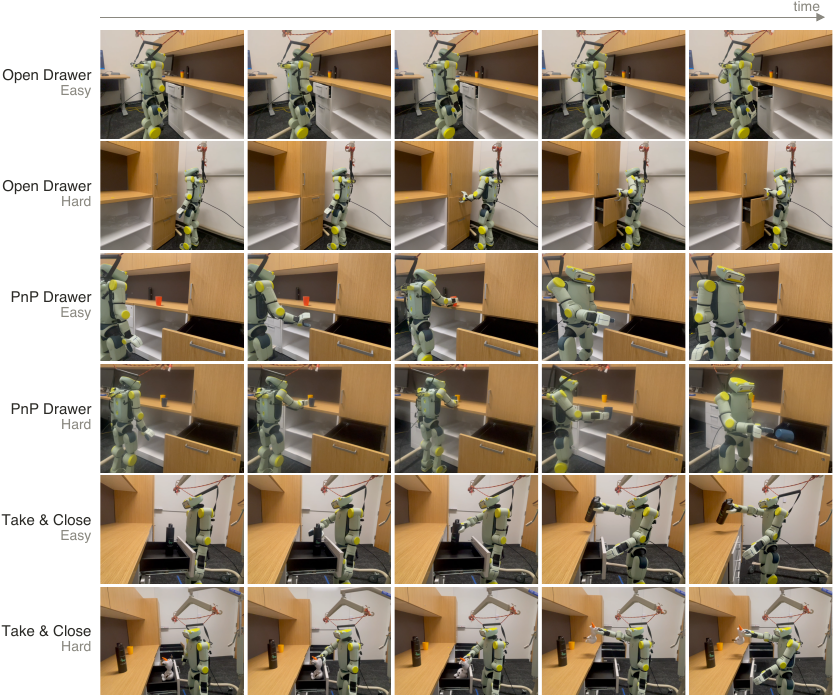}
\caption{\rev{\textbf{Real-robot rollouts of \method.} Snapshots of
context-free rollouts on the humanoid, one row per task and setting (Easy:
post-training scene; Hard: unseen drawer, unseen cup, and a stuffed toy in
place of the tumbler), with time running left to right.}}
\label{fig:real-rollouts}
\end{figure}

\rev{\paragraph{Tasks and protocol.} \emph{Open Drawer} opens a drawer.
\emph{PnP Drawer} picks up a cup and places it in an open drawer.
\emph{Take \& Close} takes a tumbler out of the drawer with one hand and
closes the drawer with the other. All checkpoints of Fig.~\ref{fig:real} and Table~\ref{tab:real},
none, within-video and \method, are post-trained with the same data and
recipe and evaluated context-free, with no human video at test time, for
$10$ trials per condition. \emph{Easy} uses the post-training scene.
\emph{Hard} changes one element never seen in post-training: a different
drawer for Open Drawer, an unseen cup for PnP Drawer, and a stuffed toy in
place of the tumbler for Take \& Close. Averaged over the Hard settings,
\method reaches $70\%$, against $57\%$ for within-video pre-training and
$43\%$ for none. Fig.~\ref{fig:real-rollouts} shows rollouts of \method in each setting.}

\begin{table}[t]
\centering
\caption{\rev{\textbf{Real humanoid, in-context, per setting.} Success rate (\%) over
$10$ trials per setting, with a human video given at test time (Fig.~\ref{fig:real}b).
Seen settings are the two post-training tasks; the others are unseen.}}
\label{tab:real-icl}
\small
\setlength{\tabcolsep}{4.5pt}
\renewcommand{\arraystretch}{1.1}
\begin{tabular}{l|ccc|ccc|cc}
    \toprule
    & \multicolumn{3}{c|}{Open Drawer} & \multicolumn{3}{c|}{PnP Drawer} & \multicolumn{2}{c}{Avg.} \\
    Pre-training & Seen & Right hand & Close & Seen & Toy grasp & Left drawer & Seen & Unseen \\
    \midrule
    None & \textbf{100} & 30 & 90 & 90 & 0 & 0 & 95.0 & 30.0 \\
    Within-video & \textbf{100} & 70 & \textbf{100} & \textbf{100} & 20 & 0 & \textbf{100.0} & 47.5 \\
    \method (Ours) & \textbf{100} & \textbf{90} & \textbf{100} & 90 & \textbf{40} & \textbf{20} & 95.0 & \textbf{62.5} \\
    \bottomrule
\end{tabular}
\end{table}

\begin{figure}[t]
\centering
\includegraphics[width=\linewidth]{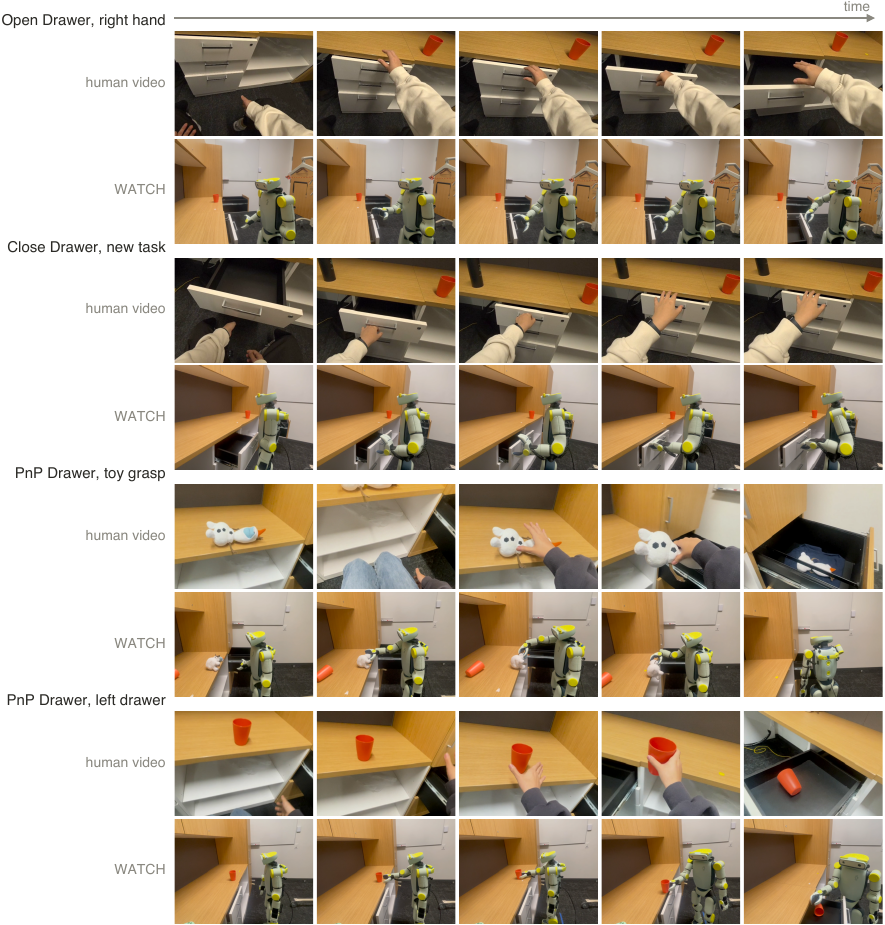}
\caption{\rev{\textbf{Real-robot in-context rollouts of \method.} For each unseen
setting, the human video given at test time (egocentric) and the \method rollout
(third person), with time running left to right.}}
\label{fig:real-rollouts-icl}
\end{figure}

\rev{\paragraph{In-context evaluation.} For in-context control, post-training
covers two tasks, opening the drawer with the left hand and placing a cup in
the right drawer. The robot data are the context-free demonstrations of these
two tasks ($20$ each, $40$ episodes; Take \& Close is left out), and each is
paired with one of $20$ human videos recorded per task, which serves as the
reference (Fig.~\ref{fig:real-icl-data}). At test time the policy receives one of three human videos recorded
for each setting and
must follow it in four settings absent from post-training: opening the drawer
with the right hand, grasping a stuffed toy with a different wrist angle,
placing the cup in the left drawer, and closing the drawer, a task not seen in
post-training. Table~\ref{tab:real-icl} reports $10$ trials per setting. All
three models solve the seen tasks; on the unseen ones, \method reaches
$62.5\%$ against $47.5\%$ for within-video pre-training and $30\%$ for none,
and it is the only model to succeed in the left drawer.
Fig.~\ref{fig:real-rollouts-icl} shows the human videos and the \method
rollouts.}

\subsection{\rev{Controlled Comparison with a Cross-Demonstration Objective}}
\label{app:v2r-ctrl}

\begin{table}[t]
\centering
\caption{\rev{\textbf{Controlled comparison of cross-demonstration pre-training
objectives.} Success rate (\%) on the Cosmos Policy backbone. Every row uses
the same human pre-training pairs, backbone, schedule, and post-training
recipe as Table~\ref{tab:ablation}; only the pre-training objective differs.}}
\label{tab:v2r-ctrl}
\small
\setlength{\tabcolsep}{5pt}
\renewcommand{\arraystretch}{1.1}
\begin{tabular}{l|cccc|ccc}
    \toprule
    & \multicolumn{4}{c|}{MimicDroid, in-context} & \multicolumn{3}{c}{TableTop-Sim, context-free} \\
    Pre-training & L1 & L2 & L3 & Avg. & Easy & Hard & Avg. \\
    \midrule
    None                    & 75.0 & 29.4 & 18.3 & 40.9 & 69.6 & 49.6 & 59.6 \\
    Within-video            & 85.8 & 44.4 & 23.3 & 51.2 & 71.6 & 51.6 & 61.6 \\
    Vid2Robot objective     & 74.2 & 41.0 & \textbf{44.1} & 53.1 & 66.0 & 38.4 & 52.2 \\
    \method                 & \textbf{91.7} & \textbf{72.5} & 40.8 & \textbf{68.3} & \textbf{76.8} & \textbf{54.0} & \textbf{65.4} \\
    \bottomrule
\end{tabular}
\end{table}

\rev{The comparison with Vid2Robot in Table~\ref{tab:icl-main} evaluates each
system as released. To isolate the pre-training formulation, we implement the
Vid2Robot training objective inside our framework and keep the backbone, the
human pre-training pairs, the $8$K-iteration schedule, and the downstream
protocol fixed. Following Vid2Robot, the model predicts the actions of the
query demonstration conditioned on its paired demonstration, with Vid2Robot's
auxiliary alignment losses between the two demonstrations and no video
prediction. Table~\ref{tab:v2r-ctrl} shows the results. In context, this
objective transfers to unseen layouts and is on par with \method on L3
($44.1$ against $40.8\%$). Context-free, it falls below no pre-training on
TableTop-Sim ($52.2$ against $59.6\%$), mostly on the Hard split ($38.4$
against $49.6\%$), while \method improves to $65.4\%$. Aligning demonstrations
and predicting actions thus yields a representation tied to the reference,
whereas predicting the paired demonstration's future video and actions yields
one that also serves policies with no reference at test time.}

\clearpage
\section{Qualitative Results: Robot Rollouts}
\label{app:rollouts}

\subsection{MimicDroid, In-Context}
\label{app:rollouts-icl}

\begin{figure}[p]
    \centering
    \begin{subfigure}{\linewidth}
        \centering
        \includegraphics[width=\linewidth]{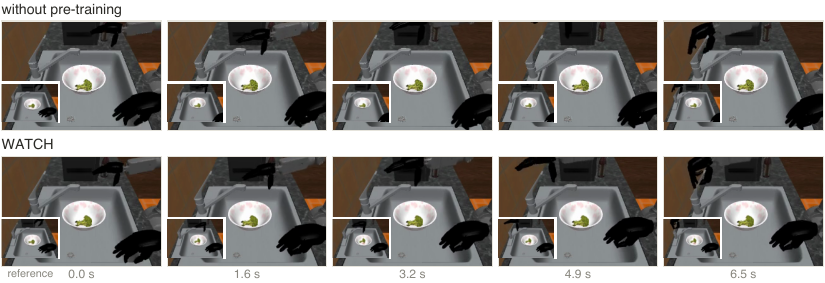}
        \caption{\texttt{TurnOnFaucet} (L3)}
    \end{subfigure}\\[0pt]
    \begin{subfigure}{\linewidth}
        \centering
        \includegraphics[width=\linewidth]{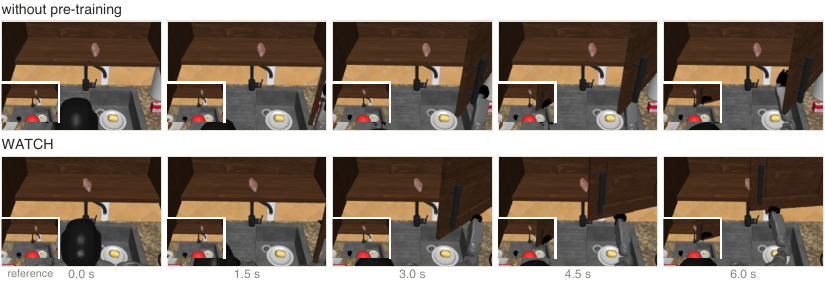}
        \caption{\texttt{CloseRightCabinetDoor} (L2)}
    \end{subfigure}\\[0pt]
    \begin{subfigure}{\linewidth}
        \centering
        \includegraphics[width=\linewidth]{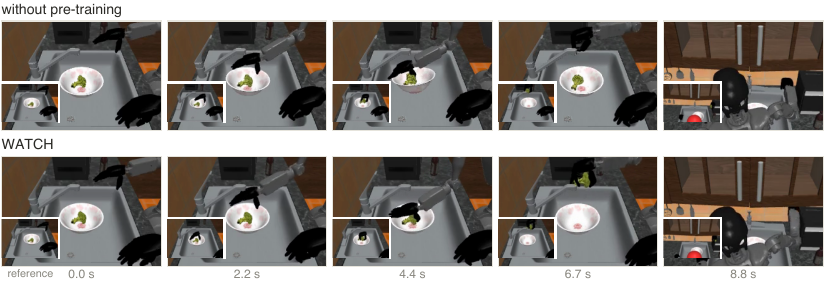}
        \caption{\texttt{PnPSinkToMicrowaveTop} (L3)}
    \end{subfigure}\\[0pt]
    \begin{subfigure}{\linewidth}
        \centering
        \includegraphics[width=\linewidth]{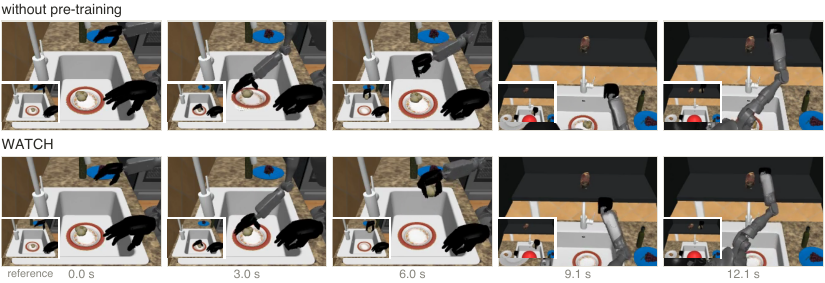}
        \caption{\texttt{PnPSinkToCabinet} (L2)}
    \end{subfigure}
    \caption{\textbf{In-context rollouts on MimicDroid.} Per task, Cosmos
    Policy post-trained in context without pre-training (top) and \method
    (bottom) at the same five time steps over the \method episode; the
    failing episode runs on to the time limit. The inset is the retrieved
    reference frame the policy is conditioned on. Each frame uses whichever
    of the front, left and right cameras shows the motion best; in (b) the two
    rows start from different initial states.}
    \label{fig:rollout-icl}
\end{figure}

In Fig.~\ref{fig:rollout-icl} the inset advances with the robot under
\method, as each chunk retrieves the reference window matching its progress,
and the hand tracks the demonstrated path; without pre-training it hovers or
stalls after the grasp.

\subsection{MimicDroid, In-Context with $2\times$ Speed Demonstrations}
\label{app:rollouts-2x}

\begin{figure}[p]
    \centering
    \begin{subfigure}{\linewidth}
        \centering
        \includegraphics[width=\linewidth]{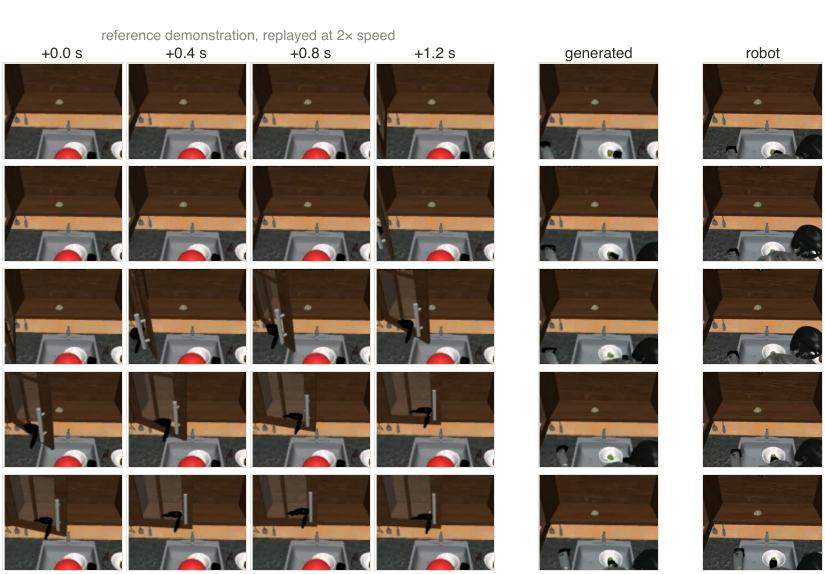}
        \caption{Cosmos Policy without pre-training}
    \end{subfigure}\\[3pt]
    \begin{subfigure}{\linewidth}
        \centering
        \includegraphics[width=\linewidth]{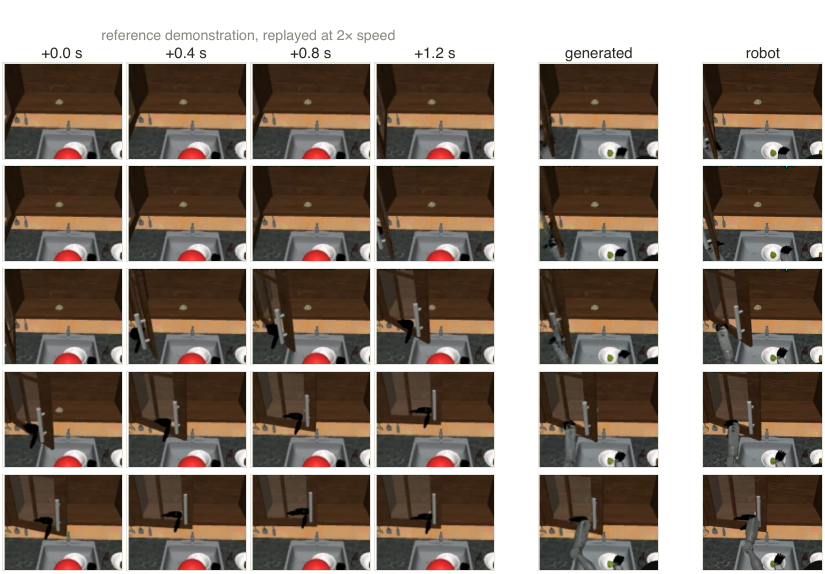}
        \caption{\method}
    \end{subfigure}
    \caption{\textbf{Reading a $2\times$-speed demonstration, chunk by chunk.}
    The same {\small\texttt{CloseLeftCabinetDoor}} episode (L3, camera
    behind the robot) under (a) Cosmos Policy without pre-training, which
    fails, and (b) \method, which succeeds. Rows are $1.6$-second action
    chunks from $1.6$ to $9.6$ s: four reference frames at $0.4$-s steps,
    replayed at $2\times$; \emph{generated}, the frame predicted for the
    chunk; \emph{robot}, the frame reached at its end. The demonstrated hand
    closes the door within a chunk; without pre-training the generated frames
    never show it closing and the arm stalls, \method spreads the motion over
    its own chunks and finishes.}
    \label{fig:rollout2x}
\end{figure}

\begin{figure}[p]
    \centering
    \begin{subfigure}{\linewidth}
        \centering
        \includegraphics[width=\linewidth]{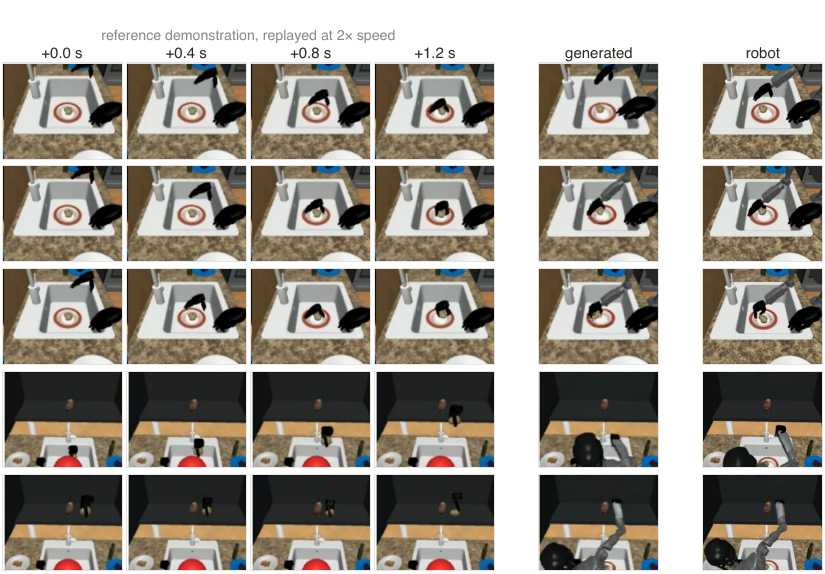}
        \caption{Cosmos Policy without pre-training}
    \end{subfigure}\\[3pt]
    \begin{subfigure}{\linewidth}
        \centering
        \includegraphics[width=\linewidth]{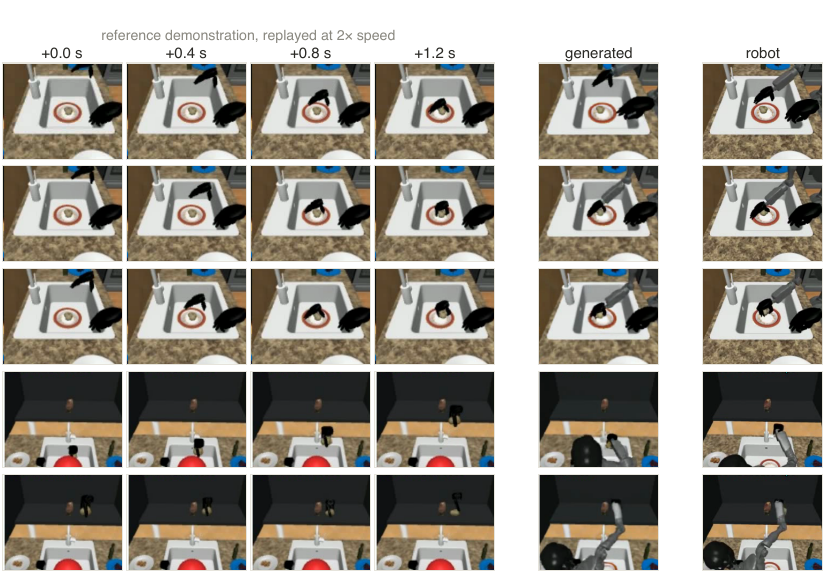}
        \caption{\method}
    \end{subfigure}
    \caption{\textbf{A two-stage task under a $2\times$-speed demonstration.}
    The same {\small\texttt{PnPSinkToCabinet}} episode (L2) under (a) Cosmos
    Policy without pre-training, which fails, and (b) \method, which
    succeeds; rows as in Fig.~\ref{fig:rollout2x}: chunks $1$--$3$ ($0$ to
    $4.8$ s) from the left camera, where the grasp is visible, then chunks $5$
    and $7$ from the camera behind the robot, which shows the transport. Both
    grasp the object; \method places it in the cabinet, whereas without
    pre-training the arm rises toward the cabinet and stays there until the
    episode times out.}
    \label{fig:rollout2x-pnp}
\end{figure}

Figs.~\ref{fig:rollout2x} and~\ref{fig:rollout2x-pnp} show demonstrations
the arm cannot copy: at $2\times$ speed the hand closes the door, or reaches
the cabinet, within one chunk. Without pre-training the generated frames stay
at the reach and the rollout fails; under \method they follow the
demonstration's progress, not its timing.

\subsection{TableTop-Sim, Context-Free}
\label{app:rollouts-tt}

\begin{figure}[!t]
    \centering
    \begin{subfigure}{\linewidth}
        \centering
        \includegraphics[width=\linewidth]{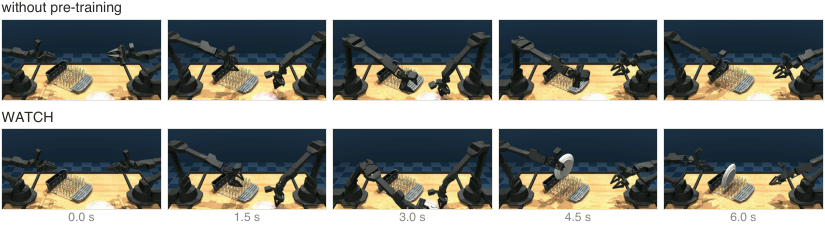}
        \caption{Dish drainer (Easy)}
    \end{subfigure}\\[0pt]
    \begin{subfigure}{\linewidth}
        \centering
        \includegraphics[width=\linewidth]{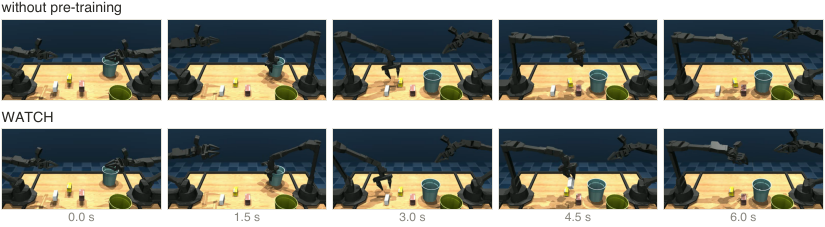}
        \caption{Box into pot (Hard)}
    \end{subfigure}\\[0pt]
    \begin{subfigure}{\linewidth}
        \centering
        \includegraphics[width=\linewidth]{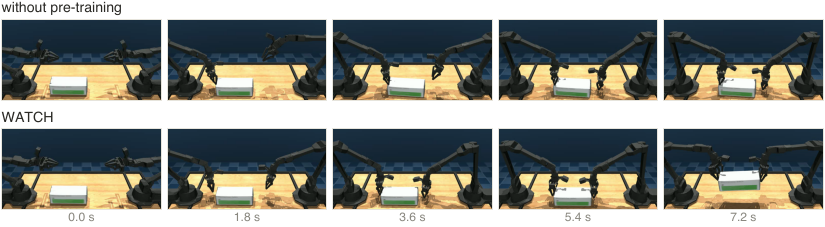}
        \caption{Lift box (Easy)}
    \end{subfigure}\\[0pt]
    \begin{subfigure}{\linewidth}
        \centering
        \includegraphics[width=\linewidth]{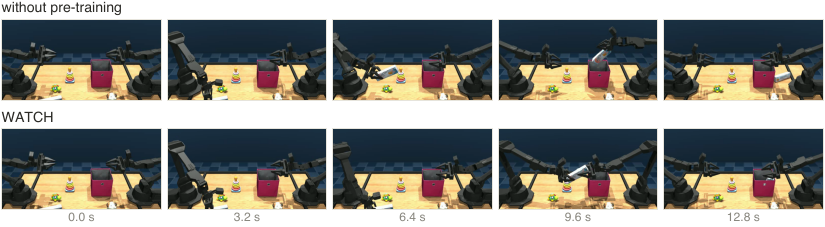}
        \caption{Handover box (Hard)}
    \end{subfigure}\\[0pt]
    \begin{subfigure}{\linewidth}
        \centering
        \includegraphics[width=\linewidth]{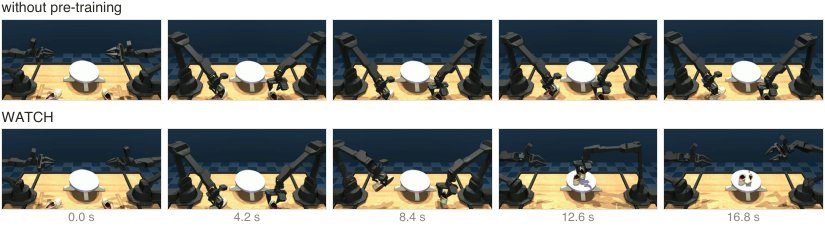}
        \caption{Shoes table (Hard)}
    \end{subfigure}
    \caption{\textbf{Context-free rollouts on TableTop-Sim.} Per task, the
    same episode under Cosmos Policy without pre-training (top) and \method
    (bottom) at the same time steps over the \method episode; the failing
    episode runs on to the time limit. Without pre-training, the plate is
    lifted and dropped (a), the small box reached but not lifted (b), the arms
    hover around the box (c), the box is dropped before the pass (d), and the
    shoes are carried to the table but not released (e).}
    \label{fig:rollout-tt}
\end{figure}

In Fig.~\ref{fig:rollout-tt}, without pre-training the policy reaches or
lifts the object but fails at the completing step; \method finishes within
the same window on both splits.

\vfill
\clearpage

\end{document}